%% file: neurips_2025.tex
\documentclass{article}

\usepackage[preprint]{neurips_2025}

\usepackage[utf8]{inputenc} 
\usepackage[T1]{fontenc}    
\usepackage{hyperref}       
\usepackage{url}            
\usepackage{booktabs}       
\usepackage{amsfonts}       
\usepackage{amssymb}
\usepackage{nicefrac}       
\usepackage{microtype}      
\usepackage{xcolor}         
\usepackage{pifont}
\usepackage{graphicx}

\usepackage{multirow}
\usepackage{wrapfig}
\usepackage{colortbl}  
\usepackage{xspace}
\usepackage{todonotes}
\usepackage[most]{tcolorbox}
\usepackage[frozencache,cachedir=.]{minted}
\tcbuselibrary{minted}

\definecolor{lightblue}{rgb}{108, 97, 250}

\newcommand{\TimeSeriesGym}{\texttt{TimeSeriesGym}\xspace}
\newcommand{\AIDE}{\texttt{AIDE}\xspace}
\newcommand{\OpenHands}{\texttt{OpenHands}\xspace}
\newcommand{\cmark}{\textcolor{green!60!black}{\ding{51}}} 
\newcommand{\xmark}{\textcolor{red}{\ding{55}}} 
\newcommand{\challenge}[1]{\textcolor{gray}{\texttt{#1}}}

\definecolor{bg}{gray}{0.95}
\DeclareTCBListing{mintedbox}{O{}m!O{}}{%
  breakable=true,
  listing engine=minted,
  listing only,
  minted language=#2,
  minted style=default,
  minted options={%
    gobble=0,
    breaklines=true,
    breakafter=,,
    fontsize=\scriptsize,
    numbersep=0pt,
    #1},
  boxsep=0pt,
  left skip=0pt,
  right skip=0pt,
  left=0pt,
  right=0pt,
  top=0pt,
  bottom=0pt,
  arc=5pt,
  leftrule=0pt,
  rightrule=0pt,
  bottomrule=0pt,
  toprule=1pt,
  colback=bg,
  colframe=blue!20!, 
  enhanced,
  #3}

\title{\TimeSeriesGym: A Scalable Benchmark for\\(Time Series) Machine Learning Engineering Agents}

\author{%
  Yifu Cai$^{*}$, \quad Xinyu Li$^{*}$, \quad Mononito Goswami$^{*}$, \quad Michał Wiliński\thanks{These coauthors contributed equally.},\\
  \textbf{Gus Welter}, \quad \textbf{Artur Dubrawski} \\
  Auton Lab, School of Computer Science\\
  Carnegie Mellon University\\
  \texttt{\{yifuc, xinyul2, mgoswami, mwilinsk, gwelter, awd\}@cs.cmu.edu} \\
  \url{https://github.com/moment-timeseries-foundation-model/TimeSeriesGym}
}

\begin{document}

\maketitle

\input{sections/00_abstract}
\input{sections/01_introduction}
\input{sections/02_related_work}
\input{sections/03_benchmark}
\input{sections/04_experiments_and_results}
\input{sections/05_discussion}

\input{sections/06_conclusion}

\bibliographystyle{plain}
\bibliography{reference}


\section*{Reproducibility statement}
We provide \TimeSeriesGym as an open-source project under the permissive MIT License: \url{https://github.com/moment-timeseries-foundation-model/TimeSeriesGym}. The repository includes detailed documentation on running experiments, adding new challenges, and incorporating different agentic scaffolds. Tab.~\ref{tab:timeseriesgym_challenges} lists all challenges in \TimeSeriesGym, while Tab~\ref{tab:time_series_sources} provides their sources and licenses. We describe our exact experimental settings and compute resources in Sec.~\ref{sec:experiments_and_results}, with scaffold hyperparameters detailed in Tab.~\ref{tab:scaffold-hyperparams}. The cost to run each experiment is reported in Tab.~\ref{tab:average_costs}.

\newpage


\appendix
\input{sections/AB_appendix}


\newpage

\end{document}

%% file: sections/00_abstract.tex
\begin{abstract}
We introduce \TimeSeriesGym, a scalable benchmarking framework for evaluating Artificial Intelligence (AI) agents on time series machine learning engineering challenges. Existing benchmarks lack scalability, focus narrowly on model building in well-defined settings, and evaluate only a limited set of research artifacts (e.g., CSV submission files). To make AI agent benchmarking more relevant to the practice of machine learning engineering, our framework scales along two critical dimensions. First, recognizing that effective ML engineering requires a range of diverse skills, \TimeSeriesGym incorporates challenges from diverse sources 
spanning multiple domains and tasks. We design challenges to evaluate both isolated capabilities (including data handling, understanding research repositories, and code translation) and their combinations, and rather than addressing each challenge independently, we develop tools that support designing multiple challenges at scale. Second, we implement evaluation mechanisms for multiple research artifacts, including submission files, code, and models, using both precise numeric measures and more flexible LLM-based evaluation approaches. This dual strategy balances objective assessment with contextual judgment. Although our initial focus is on time series applications, our framework can be readily extended to other data modalities, broadly enhancing the comprehensiveness and practical utility of agentic AI evaluation. We open-source our benchmarking framework to facilitate future research on the ML engineering capabilities of AI agents.
\end{abstract}

%% file: sections/01_introduction.tex
\section{Introduction}
\label{sec:introduction}

AI agents~\cite{huang2024mlagentbench, jiang2025aide} have shown growing promise in automating machine learning (ML) and data science (DS) workflows. Fueled by powerful large language models (LLMs), they can reason about context, adapt to diverse tasks, and iteratively refine their solutions over long horizons. Such capabilities have the potential to significantly reduce mundane, mostly manual efforts required in ML engineering and improve the overall productivity of ML practice. To measure progress in this area, several benchmarks~\cite{huang2024mlagentbench, chan2025mlebench, jing2025dsbench, nathani2025mlgymnewframeworkbenchmark, wijk2024re, bogin-etal-2024-super, tang2023ml, padigela2025ml} have been introduced to evaluate the capabilities of AI agents on ML and DS tasks.

However, existing benchmarks have important limitations. First, many of them source ML challenges primarily from popular competitions such as Kaggle, which are well-structured and do not fully capture the complexity of real-world ML tasks.
Second, evaluations are typically outcome-based, focusing on overall task completion or eventual model performance metrics such as accuracy, while combining and obfuscating the impact of multiple 
component skills that jointly determine the outcomes, such as effective data wrangling or code quality improvement capabilities.  
Third, current benchmarks lack scalability, as tasks have to be manually curated and cannot be developed at scale.  

To enable efficient evaluation of AI agents in realistic ML scenarios, we propose \TimeSeriesGym, a scalable and agent-agnostic benchmarking framework for evaluating AI agents on time series ML engineering tasks. \TimeSeriesGym currently consists of 34 challenges that span 8 unique time series problems (forecasting, classification, time series understanding), from more than 15 domains (healthcare, finance, epidemiology). Our benchmark covers both Kaggle-style challenges and original tasks carefully designed based on real-world ML engineering practice. While we focus on time series analytics due to its prevalence in applications and under-representation in existing agent benchmarks, our framework is modality-agnostic in principle and can be easily extended to handling other data modalities (e.g., images, text, audio) via the accompanying tools for scalable task generation. \TimeSeriesGym provides an interactive \texttt{gym} environment compatible with various types of agent scaffolding, allowing seamless evaluation of agents of different types and collection of their trajectories. Beyond benchmarking, this allows \TimeSeriesGym to also serve as a data flywheel for future agent improvement through post-training using the collected trajectory data.

\begin{figure}
    \centering
    \includegraphics[width=\linewidth, trim={1pt 56pt 5pt 5pt}, clip]{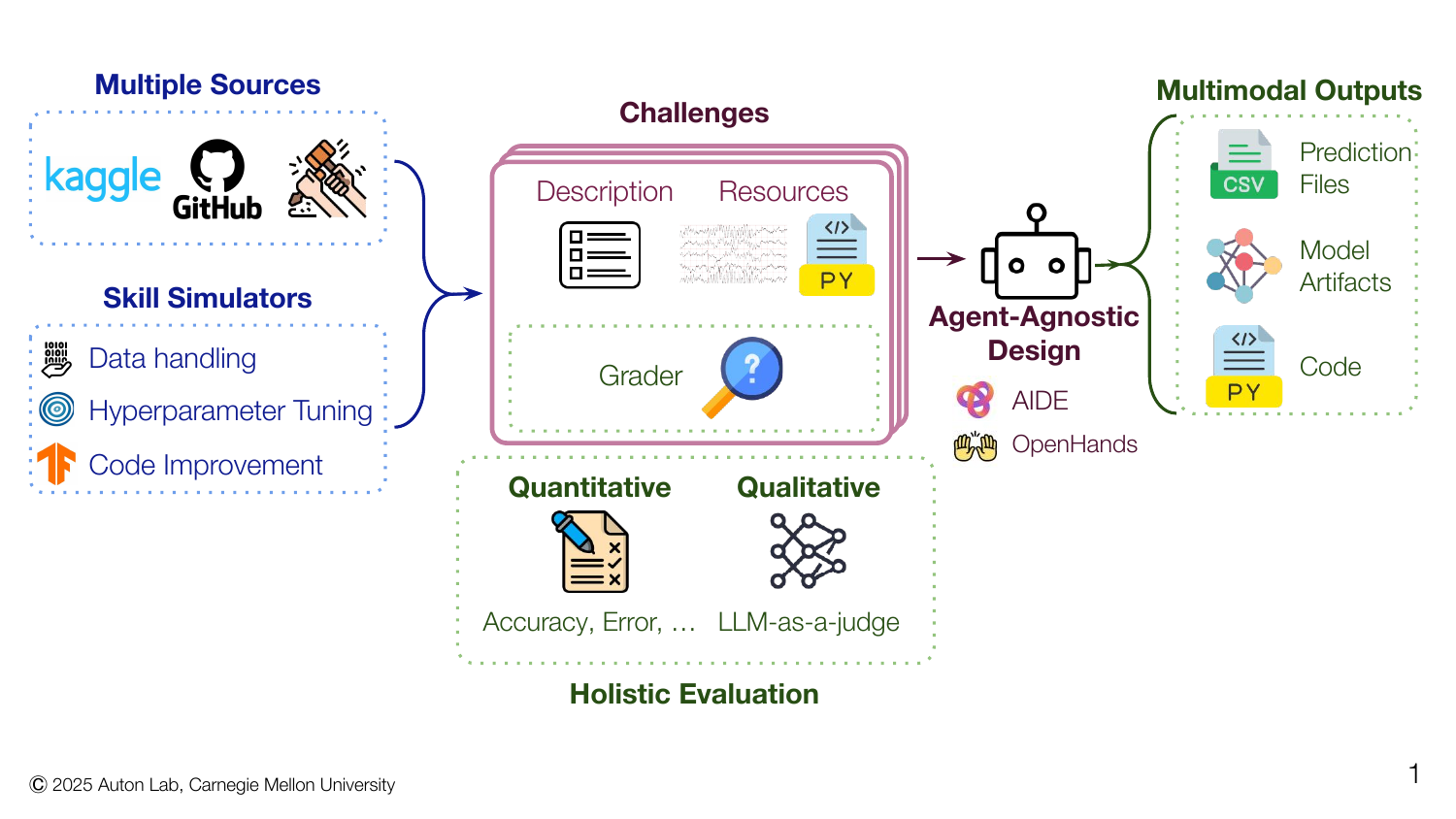}
    \caption{\TimeSeriesGym is a scalable benchmarking environment for ML engineering agents. It currently features 34 time series challenges across 8 unique time series problems, spanning more than 15 domains. 
    Challenges are either carefully designed based on real-world ML practice, 
    or sourced from Kaggle competitions and GitHub repositories. \TimeSeriesGym includes key mechanisms to enable efficient and scalable generation of new challenges. Our evaluation methodology combines precise quantitative metrics with flexible qualitative assessment, and provides specialized tools to grade various artifacts generated during ML engineering. \TimeSeriesGym is compatible with many different agent types, even those with fundamentally distinct designs.}
    \label{fig:overview}
\end{figure}

Our contributions include:
\begin{enumerate}
    \item We propose \TimeSeriesGym, the first open source benchmark environment for evaluating AI agents on time series ML engineering tasks (Fig.~\ref{fig:overview}).
    \item \TimeSeriesGym offers a scalable task generation mechanism that reduces manual efforts in task curation and ensures long-term sustainability of the benchmark.
    \item \TimeSeriesGym provides a comprehensive framework that evaluates \textbf{multimodal agent outputs} (e.g., prediction files, models, code) across \textbf{specific ML skills} (e.g., data handling, model improvement), using a \textbf{holistic approach} that combines quantitative metrics (e.g., accuracy) with qualitative assessment (e.g. LLM-as-a-judge for code utility).
\end{enumerate}

%% file: sections/02_related_work.tex
\section{Related Work}

\vspace*{-3pt}
\paragraph{Machine learning agent benchmarks.}
Several benchmarks have been proposed to evaluate LLM agents on automating ML and DS tasks. These benchmarks are typically structured around three key components: (1) task curation, (2) agent capabilities being evaluated, and (3) evaluation protocol. 
Benchmarks differ in how they curate ML/DS tasks. For example, MLE-bench~\cite{chan2025mlebench} and DSBench~\cite{jing2025dsbench} compile tasks from online competition platforms such as Kaggle, while other benchmarks source tasks from ML-related Github repositories~\cite{bogin-etal-2024-super, tang2023ml} or hand-craft tasks based on ML research problems or engineering workflows~\cite{huang2024mlagentbench, nathani2025mlgymnewframeworkbenchmark, wijk2024re, padigela2025ml}. 
With regard to agent capabilities, some benchmarks~\cite{chan2025mlebench, huang2024mlagentbench, nathani2025mlgymnewframeworkbenchmark, wijk2024re, jing2025dsbench} focus on comprehensive ML science skills by evaluating agents on end-to-end problem solving skills, while others~\cite{bogin-etal-2024-super, tang2023ml, padigela2025ml} focus on more modular engineering-oriented capabilities within the ML pipeline, such as using GitHub repositories or integrating APIs. 
Evaluation protocols also differ in output formats and granularity. MLE-bench~\cite{chan2025mlebench} and DSBench~\cite{jing2025dsbench} require agents to output results in specific formats (e.g., CSV files) that can be directly scored using predefined metrics such as accuracy, while other benchmarks~\cite{huang2024mlagentbench, nathani2025mlgymnewframeworkbenchmark, wijk2024re} allow for more flexible outputs in addition to prediction files, such as model artifacts and code. ML-Dev-Bench~\cite{padigela2025ml} further extends the evaluation by specific skills (e.g., data handling, model improvement), while SUPER~\cite{bogin-etal-2024-super} provides a more holistic evaluation by combining outcome-based evaluation with qualitative code inspection to assess agents' progress towards completing the tasks.


\begin{table}[!tb]
  \centering
  \caption{Comparison of \TimeSeriesGym with existing ML/DS agent benchmarks. Categories include \textbf{Number:} total and time series (TS) tasks in each benchmark, where each task corresponds to a unique data source (e.g., a single Kaggle competition or GitHub repository); \textbf{Source:} task origins (K: Kaggle, G: GitHub, H: Hand-crafted); \textbf{ML Capability:} coverage of ML \textbf{science} tasks (e.g., modeling, open-ended research) and \textbf{engineering} tasks (e.g., repository utilization, API integration); and \textbf{Evaluation:} capabilities for evaluating \textbf{multimodal} outputs (e.g., prediction files, model artifacts), specific ML \textbf{skills} (e.g., data handling, model improvement), and from a \textbf{holistic} perspective (combining quantitative metrics (accuracy, mean absolute error) with qualitative evaluation via code landmarks or LLM-as-a-judge approaches). We use "+" to indicate \TimeSeriesGym's scalability which enables the generation of an unlimited number of new challenges using the tools provided.
  }
  \label{tab:contribution}
  \resizebox{\textwidth}{!}{%
  \begin{tabular}{r|cc|ccc|cc|ccc}
    \toprule
     & \multicolumn{2}{c|}{\textbf{Number}} & \multicolumn{3}{c|}{\textbf{Source}} & \multicolumn{2}{c|}{\textbf{ML Capability}} & \multicolumn{3}{c}{\textbf{Evaluation}} \\
     & Total & TS & K & G & H & Science & Engineering & Multimodal & Skill-based & Holistic \\
    \midrule
    MLE-bench~\cite{chan2025mlebench} & 75 & 3 & \cmark & \xmark & \xmark  & \cmark & \xmark & \xmark & \xmark & \xmark \\
    MLAgentBench~\cite{huang2024mlagentbench} & 13 & 1 & \cmark & \xmark & \cmark & \cmark  & \xmark & \cmark & \xmark & \xmark \\
    MLGym~\cite{nathani2025mlgymnewframeworkbenchmark} & 13 & 0 & \cmark & \xmark & \cmark  & \cmark & \xmark & \cmark & \xmark & \xmark \\
    RE-Bench~\cite{wijk2024re} & 7 & 0 & \xmark & \xmark & \cmark  & \cmark & \xmark & \cmark & \xmark & \xmark \\
    DSBench\footnotemark~\cite{jing2025dsbench} & 74 & 5 & \cmark & \xmark & \xmark & \cmark & \xmark & \xmark & \xmark & \xmark \\
    SUPER\footnotemark~\cite{bogin-etal-2024-super} & 45 & 0 & \xmark & \cmark & \xmark & \xmark & \cmark & \xmark & \xmark & \cmark \\
    ML-Bench~\cite{tang2023ml} & 18 & 1 & \xmark & \cmark & \xmark  & \xmark & \cmark & \xmark & \xmark & \xmark \\
    ML-Dev-Bench~\cite{padigela2025ml} & 30 & 0 & \xmark & \xmark & \cmark  & \xmark & \cmark & \cmark & \cmark & \xmark \\
    \TimeSeriesGym (Ours) & 23+ & 23+ & \cmark & \cmark & \cmark & \cmark & \cmark & \cmark & \cmark & \cmark\\
    \bottomrule
  \end{tabular}%
  }
\end{table}
\footnotetext[1]{For DSBench, we include only data modeling tasks, while excluding data analysis tasks as they are not directly relevant to our work.}
\footnotetext[2]{For SUPER, we include repositories used to create the Expert and Masked sets of the benchmark.}

\vspace*{-3pt}
\paragraph{Scalable dynamic benchmarks and holistic evaluation.} 
Scalable benchmarks reduce manual data curation efforts by generating target problems at scale using carefully designed templates~\cite{cai2024timeseriesexam, ye2024beyond} or data engines~\cite{ho2024map}, among which TimeSeriesExam~\cite{cai2024timeseriesexam} further improves problem sample quality by applying Item Response Theory (IRT)~\cite{embretson2013item, guinet2024automated} to intelligently select questions with contextualized difficulty and appropriate discrimination. 
To remain effective against data contamination from LLM pretraining, dynamic benchmarks such as GAIA~\cite{mialon2023gaia} and LiveCodeBench~\cite{jain2024livecodebench} propose to continually incorporate problems newly released after LLM training cut-offs.
While most benchmarks target specific capabilities, holistic evaluation~\cite{liang2022holistic, goswami2023aqua} provides a comprehensive picture through evaluating models on a wide range of datasets and tasks across diverse domains using multiple complementary metrics, to capture both the breadth and depth of model capabilities.   

%% file: sections/03_benchmark.tex
\section{TimeSeriesGym}
\label{sec:timeseriegym}


\TimeSeriesGym is envisioned as a scalable benchmarking environment for time series machine learning engineering. The current version features 34 challenges from 23 unique data sources across 8 unique time series problems, spanning more than 15 domains. These challenges evaluate AI agents on a range of realistic ML engineering skills beyond just model development, including data labeling, model selection, and the utilization, improvement, and migration of research code (Tab.~\ref{tab:timeseriesgym_challenges}). \TimeSeriesGym also provides tools for rapidly developing new challenges to test specific skills and for evaluating the diverse artifacts commonly generated during ML engineering processes.

Each challenge in TimeSeriesGym is organized with a consistent structure:
(1) \textbf{resources} including datasets, code repositories, related paper(s) and documentation relevant to the challenge;
(2) a \textbf{description file} that outlines the challenge parameters, available resources, and provides specific instructions and hints for successful completion; and
(3) \textbf{challenge-specific grading functions} to evaluate agent submissions. Some challenges also include leaderboards to rank agent submissions against human performance. These leaderboards are readily available for, e.g., challenges derived from Kaggle competitions.

\textbf{The challenges in TimeSeriesGym} are derived from Kaggle competitions (currently, n = 12) and popular benchmarks and research code repositories for time series modeling (\textit{\TimeSeriesGym Originals}, n = 14).
We prioritized challenges that reflect core skills that are regularly exercised by ML engineers, researchers, and data scientists.

Each challenge is specifically chosen or designed to evaluate one or more of the following skills: 
(1)~\underline{\textit{Data Handling:}} Ability to handle missing data, use data labeling tools, and leverage multi-source data for model building.
(2) \underline{\textit{Modeling:}} Ability to develop useful time-series ML models, tune hyperparameters, perform model selection, and understand, utilize, migrate and improve the quality of research code.
(3) \underline{\textit{Benchmarking:}} Training and rigorously evaluating ML models using standard benchmarks. 
In selecting these challenges, we aimed at a broad coverage across diverse domains (e.g., healthcare, finance, epidemiology) and time series problems (forecasting, classification, time series understanding). Tab.~\ref{tab:timeseriesgym_challenges} provides a comprehensive overview of each challenge within \TimeSeriesGym, including its domain, core problem, evaluation metric, and the skills required to address it.

To identify Kaggle challenges for inclusion in the \TimeSeriesGym, we began with the Meta Kaggle dataset \cite{metakaggle}, focusing specifically on Featured and Research competitions. Featured competitions are real-world ML challenges that pose difficult, commercially oriented prediction problems, while Research competitions offer opportunities to work on problems that may not have clean or straightforward solutions\footnote{\url{https://www.kaggle.com/docs/competitions}}. We employed \texttt{Gemini 2.0 Flash} to analyze competition descriptions and titles, identifying 453 competitions that likely involve time series data. Subsequently, these were ranked based on three key metrics: participant count, maximum reward offered, and presence of a public leaderboard. From the resulting shortlist of 100 high-quality competitions, we made our final selections to ensure comprehensive coverage across diverse tasks and domains within the time series analytics landscape. 

To complement the selected Kaggle challenges, we include 14 \TimeSeriesGym Original challenges, manually curated based on recommendations from experienced ML engineers and researchers. These challenges are specifically designed to evaluate advanced technical skills that Kaggle competitions typically cannot easily assess, yet are essential for effective ML engineering. Examples include utilizing state-of-the-art models (e.g., \challenge{Implement the \texttt{MOMENT}~\cite{goswami2024moment} time series foundation model for anomaly detection}), migrating frameworks (e.g., \challenge{Convert ResNet-1D classification models from TensorFlow to PyTorch}), and improving research code quality (e.g., \challenge{Improve PTB-XL ECG Classification Code}. These capabilities represent critical competencies of skilled ML engineers that extend beyond the scope of standard Kaggle-like competitions.

Running experiments on \TimeSeriesGym can be resource-intensive and costly. Therefore, we propose \texttt{TimeSeriesGym-Lite}, a carefully selected subset of six challenges designed to efficiently evaluate AI agents on critical ML engineering skills while maintaining coverage across multiple domains and time series problems. This collection enables rapid and cost-effective assessment of novel AI agents without sacrificing the diversity of skills being tested (see Tab.~\ref{tab:timeseriesgym_challenges_lite}).

\vspace*{-3pt}
\paragraph{Multimodal, skill-based, holistic evaluation.} Existing benchmarks typically summarize agent performance using metrics such as accuracy, completion rate, or competition rankings~\cite{chan2025mlebench}. Although these metrics provide useful summaries, they do not offer much actionable feedback for improvement. \TimeSeriesGym addresses this limitation through an evaluation framework designed to provide specific actionable feedback through multiple complementary approaches.
First, we design challenges that isolate and test specific skills, such as handling missing data (e.g., \challenge{Optiver Realized Volatility Prediction with Missing Data}). Poor performance on these targeted challenges clearly indicates potential skill gaps, enabling developers to focus their efforts on specific skills.
Second, we develop fine-grained evaluation tools that assess multiple dimensions of performance simultaneously. For example, in code migration tasks (e.g., \challenge{Convert ResNet from TensorFlow to PyTorch}), our evaluation tools examine whether an agent follows instructions and naming conventions, completes all required function definitions, in addition to successful execution-- providing a multidimensional performance profile rather than a binary success/failure indicator.

Our evaluation methodology deliberately combines multiple assessment approaches: quantitative metrics (accuracy, mean absolute error), programmatic analysis (regular expression matching, code inspection), and qualitative evaluation (LLM-as-a-judge) (see Appendix~\ref{app:holistic_grading}). This hybrid approach balances the reliability of objective metrics with the flexibility of subjective assessment. Although LLM-based evaluation offers valuable insight, especially for open-ended tasks such as research code enhancement, we recognize that LLMs can be inconsistent and prone to hallucination. To mitigate these concerns, we strategically complement subjective assessments with precise quantitative metrics, creating a robust evaluation system that leverages the strengths of each approach while offsetting their individual limitations.

Furthermore, \TimeSeriesGym provides specialized tools to grade diverse artifacts generated throughout the ML engineering lifecycle-- from submission files (\texttt{CSV}, \texttt{H5}, etc.) to source code (\texttt{.py}) and trained models (\texttt{.pth}, \texttt{.pkl})-- enabling comprehensive assessment of the entire development process rather than focusing solely on final outputs.

\vspace*{-3pt}
\paragraph{Generating challenges at scale.} We provide key mechanisms to enable efficient and scalable generation of new challenges. First, we offer clear and detailed documentation that explains how to add new challenges to the benchmark. Second, we provide specialized tools to create skill-specific challenges (e.g., simulating missing data) and evaluate them. Using these resources, our team successfully created several new challenges in two hours of effort, each testing specific ML engineering skills. This scalability ensures that \TimeSeriesGym can grow and adapt as time series machine learning techniques continue to advance.

\subsection{Design Choices}

\paragraph{Focus on time series tasks.} We focused on time series modeling tasks for two key reasons. First, time-series data are ubiquitous, and effectively modeling them can significantly advance critical domains such as healthcare and economics. Despite this importance, existing agentic AI benchmarks include relatively few time series challenges (Tab.~\ref{tab:contribution}). Second, compared to text and images, time series data require modest resources for storage and modeling, making \TimeSeriesGym efficient to run. Moreover, modeling time-series data remains relatively underexplored outside specialized research communities, meaning that LLMs are less likely to have encountered such data and tasks during training. This characteristic, combined with the fact that \TimeSeriesGym evaluates general machine learning skills, makes it an excellent testbed to evaluate AI agent capabilities. Moreover, the tools in this benchmark can be easily used to include non-time-series problems as well.

\vspace*{-3pt}
\paragraph{How much freedom should the agents be given?} When designing challenges for \TimeSeriesGym, we had to strike a fine balance between giving agents freedom to solve problems creatively while keeping enough structure in place to allow for a precise and fine-grained evaluation. For example, in the \challenge{PTB-XL ECG Classification with Hyper-parameter Optimization} challenge, we \textit{required} agents to use a PyTorch-based neural network and save their models, files and code before and after tuning. This allowed us to inspect models and code to check if the hyperparameters changed, and measure how these changes improved performance.

\vspace*{-3pt}
\paragraph{Agent-agnostic design.} \TimeSeriesGym is agnostic to specific agent implementations. Following \texttt{MLE-bench}~\cite{chan2025mlebench}, it is easy to add new challenges and agentic scaffolds. To illustrate this flexibility, we include \textit{latest} implementations of 3 different scaffolds, \texttt{AIDE}~\cite{jiang2025aide}, \texttt{ResearchAgent}~\cite{huang2024mlagentbench}, and \texttt{OpenHands}~\cite{wang2024openhands} with fundamentally different designs. Unlike \texttt{MLGym}~\cite{nathani2025mlgymnewframeworkbenchmark}, we do not advocate for a default agentic scaffold, as we believe that agent designs will continue to evolve and no single scaffold will work best for all ML engineering tasks.

%% file: sections/04_experiments_and_results.tex
\section{Experiments and Results}
\label{sec:experiments_and_results}
\vspace*{-3pt}
\begin{table}[!ht]
\centering
\caption{\textbf{Main Results}. Each experiment was run with 3 random seeds, with results showing mean $\pm$ standard deviation. The table compares scaffold types (\texttt{OpenHands} vs. \texttt{AIDE}), model choices (\texttt{GPT-4.1}, \texttt{o3}, \texttt{Claude 3.7}), resource allocations (4/50 to 12/150 hours/steps), and time utilization approaches. Key findings include: (1) \texttt{AIDE} outperforms \texttt{OpenHands} as a scaffold, (2) the reasoning model \texttt{o3} achieves significantly higher valid submission rates (94.4\%) than other models, (3) \texttt{Claude 3.7} produces the most reasonable submissions (38.9\%), (4) doubling time resources does not consistently improve performance, and (5) interestingly, removing step-wise reminders sometimes improves reasonable submission rates.}
\label{tab:results}
\resizebox{0.9\textwidth}{!}{
\begin{tabular}{clc|ccc}
\toprule
\multirow{3}{*}{\textbf{\texttt{Lite}}} & \multirow{3}{*}{\textbf{Model}} & \multirow{2}{*}{\textbf{Resources}} & \textbf{Valid} & \textbf{Reasonable}  \\
 & & & \textbf{Submission} & \textbf{Submission}  \\
 & & (hours / steps) & (\%) & (\%) & \\
\midrule
\rowcolor{lightgray} & \texttt{OpenHands} & & & & \\
\cmark & + \texttt{gpt-4.1-2025-04-14} & \multirow{1}{*}{4 / 50} & $44.4 \pm 19.3$ & $11.1 \pm 9.6$ & \\ \midrule
\rowcolor{lightgray} & \texttt{AIDE} & & & & \\
\midrule
\xmark & + \texttt{gpt-4.1-2025-04-14} & \multirow{1}{*}{4 / 50} & $57.3\pm 7.9$& $12.5\pm0$& \\ \midrule
\multirow{3}{*}{\cmark} & + \texttt{gpt-4.1-2025-04-14} & \multirow{3}{*}{4 / 50} & $66.7 \pm 16.7$ & $27.8 \pm 9.6$ & \\ 
 & + \texttt{o3-2025-04-16} &  & $\mathbf{94.4 \pm 9.6}$ & $33.3 \pm 0.0$ & \\ 
 & + \texttt{claude-3-7-sonnet-20250219} &  & $50.0 \pm 16.7$ & $38.9 \pm 19.3$ & \\ \midrule
\rowcolor{lightgray} \multicolumn{6}{l}{\textit{Effect of Scaling Resources}} \\ \midrule
 &  &  4 / 50 & $66.7 \pm 16.7$ & $27.8 \pm 9.6$ & \\ 
\cmark & + \texttt{gpt-4.1-2025-04-14} &  8 / 100 & $72.2 \pm 9.6$ & $22.2 \pm 9.6$ & \\ 
 &  & 12 / 150 & $61.1 \pm 9.6$ & $\mathbf{50.0 \pm 0.0}$ & \\ \midrule
\rowcolor{lightgray} \multicolumn{6}{l}{\textit{Effective Utilization of Time}} \\ \midrule
\multirow{2}{*}{\cmark} & \texttt{Step-wise reminder} &  \multirow{2}{*}{4 / 50} & $66.7 \pm 16.7$ & $27.8 \pm 9.6$ & \\ 
 & \texttt{No reminder} &   & $55.6 \pm 9.6$ & $33.3 \pm 0.0$ & \\ 
\bottomrule
\end{tabular}
}
\end{table}

\vspace*{-3pt}
\paragraph{Setting.} We run agents in an Ubuntu 20.04 Docker container with all necessary resources (datasets, code repositories, etc.) and basic Python packages useful for ML engineering. Agents can access the internet and install additional packages as needed. For each challenge, agents have a maximum of 4 hours and 50 steps \cite{huang2024mlagentbench, padigela2025ml, nathani2025mlgymnewframeworkbenchmark} and use a machine with 128 vCPUs, 503 GB RAM, 1.8 TiB SSD, and a single NVIDIA A100-SXM4-80GB GPU\footnote{In practice, agents share this machine as we run multiple challenges in parallel. This represents a realistic setting similar to how ML engineers routinely share computing resources. We found no instances where this sharing might have disadvantaged any agent.}. Unless otherwise specified, we repeat each experiment with 3 different seeds (0, 1, and 2) to calculate mean and standard deviation. 


\vspace*{-3pt}
\paragraph{Cost.} On average, it cost us USD 63.00 to run \texttt{AIDE} with \texttt{gpt-4.1-2025-04-14} for a maximum of 4 hours and 50 steps on \TimeSeriesGym. In contrast, the \texttt{Lite} benchmark was much more affordable at USD 8.00
per run. Therefore, to save both time and money, we conducted most of our experiments on \TimeSeriesGym\texttt{-Lite}.

\vspace*{-3pt}
\paragraph{Metrics.} We report the raw scores achieved by each agent on every challenge (Tab~\ref{tab:timeseriesgym_challenges}). Although these scores are useful for tracking progress on individual challenges, they cannot be easily combined across different challenges. To measure the performance of agents at a high level, we report two key metrics: the percentage of challenges where the agent made a (1) \textit{valid}, and (2) \textit{reasonable} submission. A submission is valid if the grader returns any non-null score. What counts as a reasonable attempt varies by challenge type (Tab.~\ref{tab:results}). For Kaggle challenges, we define a reasonable attempt as scoring above median on the competition's public leaderboard. For the remaining challenges, a reasonable submission shows a genuine attempt at generating a valid submission, rather than hallucinating an output that matches the submission format. We determine this by examining agent logs either manually or using LLM-as-a-judge.

\subsection{Observations}

\vspace*{-3pt}
\paragraph{\texttt{AIDE} is the better open-source scaffold.} We evaluated \texttt{GPT-4.1} (\texttt{gpt-4.1-2025-04-14}) using two open-source scaffolds: \texttt{AIDE}~\cite{jiang2025aide} and \texttt{OpenHands}~\cite{wang2024openhands}. Following \texttt{MLE-bench}, we make minor modifications to adapt these scaffolds to our benchmark (see Appendix~\ref{app:implementation_details_of_scaffolds}). Our results in Tab.~\ref{tab:results} confirm previous findings: \texttt{GPT-4.1} produces more valid (66.7\% vs 44.4\%) and reasonable (27.8\% vs 11.1\%) submissions with \texttt{AIDE} than with \texttt{OpenHands}. This is expected because \texttt{AIDE} is specifically designed for data science tasks, which account for the majority of the \TimeSeriesGym challenges. 


\vspace*{-3pt}
\paragraph{Reasoning models produce substantially more valid submissions.} To identify the best base model, we conducted experiments using the best scaffold (\texttt{AIDE}) with two state-of-the-art proprietary LLMs: \texttt{GPT-4.1} (\texttt{gpt-4.1-2025-04-14}) and \texttt{Claude 3.7 Sonnet} (\texttt{claude-3-7-sonnet-20250219}), and a reasoning model \texttt{o3} (\texttt{o3-2025-04-16}). As shown in Tab.~\ref{tab:results}, our experiments on \texttt{TimeSeriesGym-Lite} revealed that \texttt{o3} created significantly more valid submissions than other models, while \texttt{Claude 3.7} produced the highest number of reasonable attempts (38.9\%). 

\vspace*{-3pt}
\paragraph{Challenges in \TimeSeriesGym are hard for state-of-the-art agents.} We tested \texttt{AIDE} with \texttt{GPT-4.1} on all \TimeSeriesGym challenges and found poor overall performance. The agent produced valid submissions for only 57.3\% of challenges and reasonable submissions for just 12.5\%. We found that this agent especially struggled with \TimeSeriesGym original challenges, where it failed to produce valid submissions for 5 out of 13 challenges (Tab.~\ref{tab:main_result_raw}). These results show that even the best agents struggle with ML engineering tasks, particularly those that go beyond standard Kaggle data science challenges and involve working with multi-file code repositories.

\vspace*{-3pt}
\paragraph{Agents do not improve with more time.} We wondered if the agents perform poorly on \TimeSeriesGym simply because they need more time. To test this idea, we ran \texttt{AIDE} with \texttt{GPT-4.1} on \TimeSeriesGym\texttt{-Lite} and gave it 2 or 3 times more hours and steps to solve each challenge. Our results show that extra time does not always improve performance (Tab.~\ref{tab:ablation}). Even with the maximum time (12 hours and 150 steps), the agent only made reasonable submissions in about 50\% of challenges. Although this might seem promising, it is not very impressive, since the bar for a ``reasonable" submission is quite low.

\begin{wrapfigure}{r}{0.4\textwidth}
    \centering
    \includegraphics[width=0.4\textwidth]{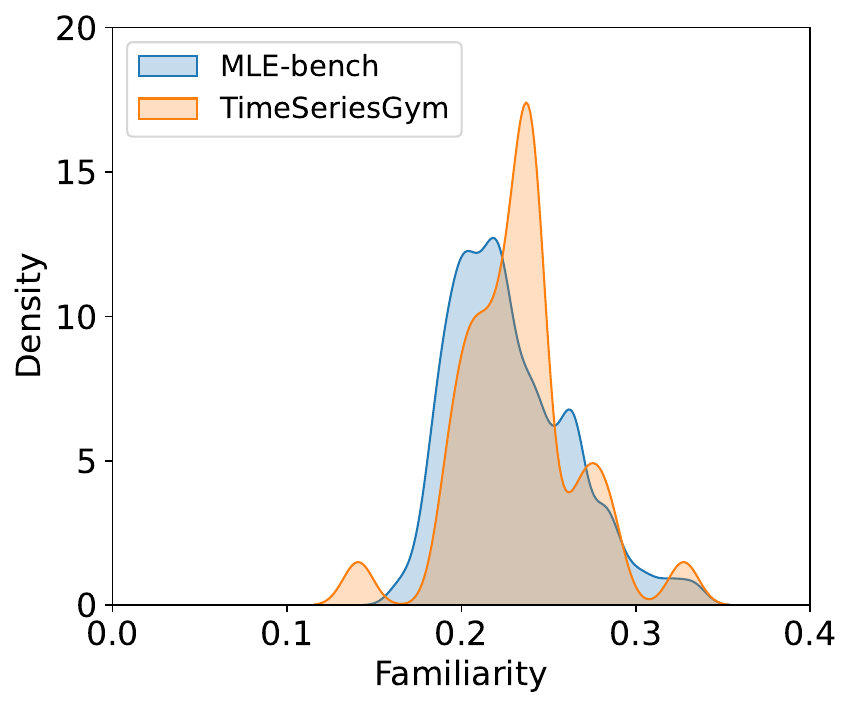}
    \caption{\texttt{GPT-4.1}'s familiarity with \TimeSeriesGym challenges, compared to its familiarity with \texttt{MLE-bench}. }
    \label{fig:familiarity}
    \vspace{-2em}
\end{wrapfigure}

\vspace*{-3pt}
\paragraph{Agents do not utilize time effectively.} We suspected that agents do not improve with more time because they do not use it well. To test this idea, we compared two settings: the default approach of reminding the agent about remaining time (and steps) before each step, versus removing these reminders completely. Surprisingly, we did not find significant differences between these settings. In fact, agents without time reminders produced more reasonable submissions. This may suggest that agents do not use their time wisely-- they tend to rush toward solutions instead of carefully exploring promising options, especially towards the end of the experiment. This raises important research questions about how to design agents that use their time and resources more strategically. 

\vspace*{-3pt}
\paragraph{Frontier LLM challenges.} Since frontier LLMs are pretrained on large-scale public data, there is a risk that they may have encountered and memorized content from public challenges, e.g., online Kaggle competition discussions or solutions, which can potentially inflate benchmark performance and limit its generalizability. 
To assess this risk, we followed the approach used by \texttt{MLE-bench} to measure \texttt{GPT-4.1}'s familiarity with \TimeSeriesGym challenges and compared the familiarity score distribution to that of \texttt{MLE-bench}. 
As shown in Fig.~\ref{fig:familiarity}, \texttt{GPT-4.1} exhibited a similar level of familiarity with \TimeSeriesGym challenges as with \texttt{MLE-bench} challenges (with a Kolmogorov–Smirnov (KS) Test~\cite{massey1951kolmogorov} p-value of 0.363, indicating no significant difference). Given that \texttt{MLE-bench} found no systematic impact of LLM familiarity on experiment results, \texttt{GPT-4.1}'s familiarity with \TimeSeriesGym is within a reasonable range and does not compromise its integrity.

\vspace*{-3pt}
\paragraph{Summary.}
This section provides a focused illustration of how \TimeSeriesGym enables efficient and cost-effective experimentation with AI agents, helping researchers uncover actionable insights about agent capabilities and limitations. Our findings demonstrate the \TimeSeriesGym's value for advancing generic ML engineering agents.

%% file: sections/05_discussion.tex
\section{Discussion, Open Questions and Opportunities}
\label{sec:discussion}

\vspace*{-3pt}
\paragraph{Key limitations of existing scaffolds.} 
Agentic scaffolds such as \texttt{AIDE} and \texttt{OpenHands} provide structured workflows that \textbf{excel in single-shot, self-contained benchmarks} (e.g., Kaggle competitions) but reveal \textbf{significant limitations in repository-level challenges} that require multiple file edits and iterative reasoning. \texttt{AIDE}'s \emph{one-step solution strategy} and \emph{fixed action set}—restricted to predefined operations such as ``data preview" when debugging—often lead to unsuccessful attempts in large codebases, as the agent's attention is diluted across irrelevant files and fails to identify critical information. Conversely, \texttt{OpenHands} supports multi-step trajectories yet suffers from a \textbf{greedy exploitation bias}: it commits fully to a single approach without exploring alternative solution paths or revisiting earlier decisions when trajectories prove unfruitful. The planning algorithm of the \texttt{CodeAct} agent used by \texttt{OpenHands} is similarly \emph{greedy and short-horizon}, limiting adaptation to complex multistage development workflows. These findings highlight the need for \textbf{more adaptive scaffolds} that dynamically expand their action repertoire, balance exploration and exploitation through parallel solution threads, and support nested workflows reflective of real-world machine learning engineering tasks. We provide illustrations of agent failures in Appendix~\ref{app:failure_mode}.

\vspace*{-3pt}
\paragraph{Data leakage and plagiarism.} In designing \TimeSeriesGym, we identify two key risks related to data leakage and plagiarism that could compromise the integrity of the benchmark: 
(1) \textbf{Pretraining contamination:} Current LLMs may have been exposed to public content from existing challenges (e.g., Kaggle competitions), including task descriptions, data, or shared solutions. This can lead to memorization and inflated performance that overstates agents' true capabilities, and (2) \textbf{Future LLM leakage:} Once the benchmark is public, future LLMs may be pretrained on its content, making the benchmark less effective in evaluating real generalization.

To address such risks, we present both empirical findings and mitigation strategies. 
For case (1), we have two key observations. 
First, in both Kaggle-based and original challenges in \TimeSeriesGym, agents either performed poorly or did not produce valid output, suggesting minimal benefit from any potential LLM contamination.
Second, we conducted a formal analysis using available tools to assess agents' familiarity with all competitions in this benchmark. The results show no evidence of systematic prior exposure or memorization, further supporting the integrity of the benchmark in its current state.
For case (2), the scalability of \TimeSeriesGym enables efficient generation of new challenges and skill-specific variations. This allows the benchmark to evolve continuously and remain effective even if the current version is eventually included in future LLM pretraining. 

Finally, we raise a broader question around plagiarism and code reuse. Several \TimeSeriesGym challenges, such as leveraging \texttt{MOMENT}~\cite{goswami2024moment} for anomaly detection, require agents to use existing code repositories to solve open-ended ML tasks. In such cases, it becomes difficult to clearly define and assess plagiarism. For example, if an agent cites the code it uses, should it be considered plagiarism or appropriate reuse, similar to how human ML practitioners build on public code with proper reference? As the ability to effectively and properly leverage existing resources is important in real-world ML practice, we believe that it is crucial to develop clear, legally correct definitions and evaluation criteria for data contamination and plagiarism in the context of LLM agents. We highlight this as an important direction for future work.

\vspace*{-3pt}
\paragraph{Defining and measuring success.} 
What does it mean for an agent to be successful?
For Kaggle tasks, while comparing an agent's performance against human leaderboards seems intuitive, it presents challenges. 
\TimeSeriesGym utilizes different training and testing splits and reimplements the grading mechanisms from the original Kaggle competitions (as original Kaggle test sets are private), making direct leaderboard comparisons potentially misleading.
Additionally, in challenges like code migration, real-world utility does not always require perfect, bug-free code: partial, buggy solutions may still accelerate development when iterated by human engineers. Thus, our current evaluation approaches have inherent limitations.

We propose several desiderata for improving the success metrics. These metrics should be rigorous and objective, yet flexible enough to preserve agent creativity and autonomy. They should also yield actionable insights, helping identify specific deficiencies and guide future improvements in agent design. In this work, we take a step in this direction by enabling skill-based and holistic evaluations, offering a more comprehensive understanding of agent capabilities and limitations. Moving forward, we believe that the development of robust, holistic and diagnostic success metrics remains an important research direction and requires community discussion.

\vspace*{-3pt}
\paragraph{Comparative characterization.} 
In the practice of ML application, one often needs to choose an appropriate approach from a short list of eligible options. For this, conducting pairwise comparisons among the alternatives often helps to assess their relative advantages and limitations. Inspired by~\cite{cai2024timeseriesexam}, we envision an automated process that iteratively emphasizes the apparent differences in performance between any pair of agents, or one versus a collection of alternative agents, so that particular strengths or vulnerabilities can be exposed, characterized, and potentially addressed.  

\vspace*{-3pt}
\paragraph{Optimal resource allocation.} Consistent with previous work, agents were given 4 hours and 50 steps to solve each challenge - but is this sufficient? Alternative frameworks like \texttt{MLE-bench} provide substantially more resources (24 hours and approximately 2000 steps). Our scaling experiments, which gave agents up to 12 hours and 150 steps for a subset of challenges, did not reveal significant performance improvements. Therefore, we believe that further increasing resources is an option, but practical academic budget constraints make such approaches largely infeasible. This raises important questions about how to balance resource limitations with fair opportunities to assess AI agents.

\vspace*{-3pt}
\paragraph{Societal impact.} AI agents promise to substantially reduce manual effort in ML engineering while expanding the productivity and accessibility of ML tools. This automation presents several social implications worth considering. First, by lowering technical barriers, these agents could democratize ML capabilities, allowing users without an extensive programming background to leverage advanced analytics. Second, automated ML workflows can accelerate scientific discovery in multiple application domains, including healthcare, climate science, and materials research. 
However, several challenges require careful attention from the community. The primary concern is proper attribution when agents repurpose existing code, potentially obscuring original authorship and violating licenses. Furthermore, automated ML systems can perpetuate or amplify existing biases in training data without human oversight. There is also the risk of workforce displacement for entry-level ML engineers as routine tasks become increasingly automated. Furthermore, these agents might generate plausible but flawed solutions that appear correct to non-experts, leading to undetected errors in critical applications. The resource-intensive nature of running sophisticated agents could also exacerbate computational divides between well-resourced and under-resourced organizations. As we advance agent capabilities through benchmarks like \TimeSeriesGym, the research community must simultaneously develop frameworks for responsible deployment that address these challenges while maximizing societal benefits.

%% file: sections/06_conclusion.tex
\section{Conclusion}

We propose \TimeSeriesGym, a scalable and agent-agnostic benchmarking framework for evaluating AI agents (with various scaffoldings) on time series ML engineering tasks. 
By curating tasks that reflect real-world ML practice from diverse sources, enabling scalable task generation, and supporting multimodal, skill-based, holistic evaluation, \TimeSeriesGym provides a practical and extensible testbed for advancing AI agents in ML engineering. 
Our experiments show that while frontier LLMs combined with \texttt{AIDE} scaffolding~\cite{jiang2025aide} can achieve moderate to high success rates in producing valid submissions, they still do not generate reasonable solutions, particularly on \textit{\TimeSeriesGym original} challenges that emulate the complexity of real-world time series tasks.
This highlights current limitations in agent capabilities to effectively understand and solve realistic time series tasks.
By open-sourcing \TimeSeriesGym, we aim to facilitate a deeper understanding of the ML engineering capabilities of AI agents, provide actionable insights on future development, and support the collection of agent interaction trajectories to drive continuous improvement of AI agents through post-training and refinement.


%% file: sections/AB_appendix.tex
\section{\TimeSeriesGym Challenges}
\label{app:timeseriesgym_challenges}

\begin{table}[htbp]
\centering
\renewcommand{\arraystretch}{1.3} 
\resizebox{\textwidth}{!}{%
\begin{tabular}{r|cccc}
\toprule
\textbf{Challenge} & \textbf{Problem} & \textbf{Domain} & \textbf{Skills} & \textbf{Evaluation Metric} \\
\midrule
\multicolumn{5}{l}{\textit{\textbf{Kaggle Challenges}}} \\
\midrule
\begin{tabular}[c]{@{}c@{}}AMP-Parkinson's Disease\\ Progression Prediction\end{tabular} & Time-to-Event Regression & Healthcare & \multirow{18}{*}{\begin{tabular}[c]{@{}c@{}}Data Handling\\(Dealing with Missing Values,\\ Utilize Multi-Source Data)\\\\ Modeling \\(Hyper-parameter Tuning \\ \& Model Selection)\end{tabular}} & \begin{tabular}[c]{@{}c@{}}Symmetric Mean Absolute\\ Percentage Error\end{tabular} \\
ASHRAE - Great Energy Predictor III & Forecasting & Energy & & \begin{tabular}[c]{@{}c@{}}Root Mean Square\\ Logarithmic Error\end{tabular} \\
Child Mind Institute-- Detect Sleep States & Classification & Healthcare & & \begin{tabular}[c]{@{}c@{}}Event Detection\\ Average Precision\end{tabular} \\
Google Brain - Ventilator Pressure Prediction & Regression & Healthcare & & Mean Absolute Error \\
G2Net Gravitational Wave Detection & Classification & Geology & & \begin{tabular}[c]{@{}c@{}}Area Under\\ ROC Curve\end{tabular} \\
HMS - Harmful Brain Activity Classification & Classification & Healthcare & & KL Divergence \\
LANL Earthquake Prediction & Time-to-Event Regression & Geology & & Mean Absolute Error \\
M5 Forecasting - Accuracy & Forecasting & Sales & & \begin{tabular}[c]{@{}c@{}}Weighted Root Mean Squared\\ Scaled Error\end{tabular} \\
Online Product Sales & Forecasting & Sales & & \begin{tabular}[c]{@{}c@{}}Root Mean Square\\ Logarithmic Error\end{tabular} \\
Optiver Realized Volatility Prediction & Forecasting & Finance & & \begin{tabular}[c]{@{}c@{}}Root Mean Square\\ Percentage Error\end{tabular} \\
Recruit Restaurant Visitor Forecasting & Forecasting & Sales & & \begin{tabular}[c]{@{}c@{}}Root Mean Square\\ Logarithmic Error\end{tabular} \\
Sberbank Russian Housing Market & Forecasting & Housing & & \begin{tabular}[c]{@{}c@{}}Root Mean Square\\ Logarithmic Error\end{tabular} \\
\midrule
\multicolumn{5}{l}{\textit{\textbf{TimeSeriesGym Originals}}} \\
\midrule
\begin{tabular}[c]{@{}r@{}}Convert ResNet TensorFlow\\implementation to PyTorch\end{tabular} & Classification & \multirow{4}{*}{Algorithm} & \multirow{4}{*}{Code Migration} & \multirow{4}{*}{Custom Code Grading} \\
\begin{tabular}[c]{@{}r@{}}Convert STOMP Algorithm\\implementation in R to Python\end{tabular} & Data Mining &  &  &  \\
\begin{tabular}[c]{@{}r@{}}Evaluate MOIRAI time series foundation model\\ on the Context Is Key (CiK) benchmark\end{tabular} & \multirow{4}{*}{Context-aided Forecasting} & \multirow{4}{*}{\begin{tabular}[c]{@{}c@{}}Climatology, Economics, Energy, \\ Mechanics,  Public Safety, Retail,\\ Synthetic, Transportation\end{tabular}} & \multirow{19}{*}{\begin{tabular}[c]{@{}c@{}}Modeling \\ (Using Research Code)\end{tabular}} &  \multirow{4}{*}{Resolved (Binary)} \\
\begin{tabular}[c]{@{}r@{}}Evaluate Chronos time series foundation model\\ on the NN5 dataset within Context Is Key\\ (CiK) benchmark\end{tabular} &  &  &  &  \\
Implement \& Evaluate CSDI to Impute PM2.5 Data & \multirow{2}{*}{Imputation} & \multirow{2}{*}{Weather} &  & \multirow{2}{*}{Mean Absolute Error} \\
Train \& Evaluate CSDI to Impute PM2.5 Data & & & & \\
\begin{tabular}[c]{@{}r@{}}GIFT-EVAL: A Benchmark for General Time Series\\ Forecasting Model Evaluation\end{tabular} & Forecasting & \begin{tabular}[c]{@{}c@{}}Nature, Web, CloudOps, \\ Economics/Finance, Energy, \\ Sales, Transportation, \end{tabular} &  & Mean Absolute Percentage Error \\
Hexagon ML UCR Time Series Anomaly Detection & Anomaly Detection & \begin{tabular}[c]{@{}c@{}}Healthcare, Gait, Energy,\\ Synthetic, Devices\end{tabular} & & \begin{tabular}[c]{@{}c@{}}Adjusted Best\\ F1 Score\end{tabular} \\
\begin{tabular}[c]{@{}r@{}}Long Horizon Time Series Forecasting\\ Using Time Series Library\end{tabular} & Forecasting & \begin{tabular}[c]{@{}c@{}}Energy, Epidemiology, Finance,\\ Transportation, Weather\end{tabular} &  & \begin{tabular}[c]{@{}c@{}}Mean Squarred\\ Error\end{tabular} \\
\begin{tabular}[c]{@{}r@{}}Long-Horizon Weather Forecasting using\\ Time Series Library's Itransformer\end{tabular} & Forecasting & Weather &  & Exact Match \\
MIT-BIH ECG Arrhythmia Detection & Classification & Healthcare &  & Accuracy \\
\begin{tabular}[c]{@{}r@{}}MOMENT for Anomaly Detection\\ on UCR datasets\end{tabular} & Anomaly Detection & \begin{tabular}[c]{@{}c@{}}Healthcare, Gait, Energy,\\ Synthetic, Devices\end{tabular} &  & Exact Match \\
PTB-XL ECG Classification & Classification & Healthcare &  & Accuracy \\
TimeSeriesExam: A Time Series Understanding Exam & Time Series Understanding & Synthetic & Time Series Understanding & Accuracy \\
\midrule
\multicolumn{5}{l}{\textit{\textbf{Derived Challenges}}} \\
\midrule
Google Brain - Ventilator Pressure Prediction & Regression & Healthcare & \begin{tabular}[c]{@{}c@{}}Data Handling\\ (Dealing with missing data)\end{tabular} & \begin{tabular}[c]{@{}c@{}}Mean Absolute\\ Error\end{tabular} \\
Improve PTB-XL ECG Classification Code & Classification & Healthcare & \begin{tabular}[c]{@{}c@{}}Code Enhancement\\(Experiment Tracking,\\Readability, Reproducibility)\end{tabular} \\
\begin{tabular}[c]{@{}r@{}}MIT-BIH Arrhythmia Detection\\with Weak Supervision\end{tabular} & Classification & Healthcare & \begin{tabular}[c]{@{}c@{}}Data Handling\\ (Labeling)\end{tabular} & Accuracy \\
Optiver Realized Volatility Prediction & Forecasting & Finance & \begin{tabular}[c]{@{}c@{}}Data Handling\\ (Dealing with missing data)\end{tabular} & \begin{tabular}[c]{@{}c@{}}Root Mean Square\\ Percentage Error\end{tabular} \\
\begin{tabular}[c]{@{}r@{}}Optiver Realized Volatility Prediction\\ with Hyper-parameter Optimization\end{tabular} & Forecasting & Finance & \multirow{4}{*}{\begin{tabular}[c]{@{}c@{}}Modeling (Hyper-parameter\\ Tuning \& Model Selection)\end{tabular}} & \begin{tabular}[c]{@{}c@{}}Improvement in Root Mean\\ Square Percentage Error\end{tabular} \\
\begin{tabular}[c]{@{}r@{}}PTB-XL ECG Classification\\ with Hyperparameter Optimization\end{tabular} & Classification & Healthcare &  & Improvement in Accuracy \\
\bottomrule
\end{tabular}%
}
\caption{This table presents the \TimeSeriesGym benchmark's diverse collection of time series challenges across three categories: Kaggle Challenges, \TimeSeriesGym Originals, and Derived Challenges. The challenges span multiple domains (healthcare, finance, energy, weather, transportation), problem types (classification, regression, forecasting, anomaly detection), and required skills (data handling, model building, code migration). Each challenge uses appropriate evaluation metrics for its task type. The benchmark combines established Kaggle competitions with novel custom tasks, creating a comprehensive testbed for evaluating ML engineering agents across realistic scenarios that practitioners face in real-world applications.}
\label{tab:timeseriesgym_challenges}
\end{table}

\begin{table}[htbp]
\centering
\renewcommand{\arraystretch}{1.3} 
\resizebox{\textwidth}{!}{%
\begin{tabular}{lll}
\toprule
\textbf{Challenge} & \textbf{Source} & \textbf{License} \\
\midrule
\multicolumn{3}{l}{\textbf{\textit{Kaggle Challenges}}} \\
\midrule
\begin{tabular}[c]{@{}l@{}}AMP-Parkinson's Disease\\ Progression Prediction\end{tabular} & 
\href{https://www.kaggle.com/competitions/amp-parkinsons-disease-progression-prediction}{Kaggle} & 
\href{https://www.kaggle.com/competitions/amp-parkinsons-disease-progression-prediction/rules\#7.-competition-data.}{Subject to Competition Rules} \\

ASHRAE - Great Energy Predictor III & 
\href{https://www.kaggle.com/competitions/ashrae-energy-prediction}{Kaggle} & 
\href{https://www.kaggle.com/competitions/ashrae-energy-prediction/rules\#7.-competition-data.}{Subject to Competition Rules} \\

\begin{tabular}[c]{@{}l@{}}Child Mind Institute--\\ Detect Sleep States\end{tabular} & 
\href{https://www.kaggle.com/competitions/child-mind-institute-detect-sleep-states/overview}{Kaggle} & 
\href{https://www.kaggle.com/competitions/child-mind-institute-detect-sleep-states/rules\#7.-competition-data.}{CC BY-NC-SA 4.0} \\

\begin{tabular}[c]{@{}l@{}}Google Brain - Ventilator\\ Pressure Prediction\end{tabular} & 
\href{https://www.kaggle.com/competitions/ventilator-pressure-prediction}{Kaggle} & 
\href{https://www.kaggle.com/competitions/ventilator-pressure-prediction/rules\#7.-competition-data.}{Subject to Competition Rules} \\

\begin{tabular}[c]{@{}l@{}}G2Net Gravitational\\ Wave Detection\end{tabular} & 
\href{https://www.kaggle.com/competitions/g2net-gravitational-wave-detection}{Kaggle} & 
\href{https://www.kaggle.com/competitions/g2net-gravitational-wave-detection/rules\#7-competition-data}{Subject to Competition Rules} \\

\begin{tabular}[c]{@{}l@{}}HMS - Harmful Brain\\ Activity Classification\end{tabular} & 
\href{https://www.kaggle.com/competitions/hms-harmful-brain-activity-classification}{Kaggle} & 
\href{https://www.kaggle.com/competitions/hms-harmful-brain-activity-classification/rules\#7.-competition-data.}{CC BY-NC 4.0} \\

LANL Earthquake Prediction & 
\href{https://www.kaggle.com/competitions/LANL-Earthquake-Prediction}{Kaggle} & 
\href{https://www.kaggle.com/competitions/LANL-Earthquake-Prediction/rules\#7-competition-data}{Subject to Competition Rules} \\

M5 Forecasting - Accuracy & 
\href{https://www.kaggle.com/competitions/m5-forecasting-accuracy}{Kaggle} & 
\href{https://www.kaggle.com/competitions/m5-forecasting-accuracy/rules\#7-competition-data}{Subject to Competition Rules} \\

Online Product Sales & 
\href{https://www.kaggle.com/c/online-sales/overview}{Kaggle} & 
\href{https://www.kaggle.com/competitions/online-sales/rules\#7-competition-data}{Subject to Competition Rules} \\

\begin{tabular}[c]{@{}l@{}}Optiver Realized\\ Volatility Prediction\end{tabular} & 
\href{https://www.kaggle.com/competitions/optiver-realized-volatility-prediction}{Kaggle} & 
\href{https://www.kaggle.com/competitions/optiver-realized-volatility-prediction/rules\#7-competition-data}{Subject to Competition Rules} \\

\begin{tabular}[c]{@{}l@{}}Recruit Restaurant\\ Visitor Forecasting\end{tabular} & 
\href{https://www.kaggle.com/c/recruit-restaurant-visitor-forecasting}{Kaggle} & 
\href{https://www.kaggle.com/competitions/recruit-restaurant-visitor-forecasting/rules\#7-competition-data}{Subject to Competition Rules} \\

\begin{tabular}[c]{@{}l@{}}Sberbank Russian\\ Housing Market\end{tabular} & 
\href{https://www.kaggle.com/competitions/sberbank-russian-housing-market/data}{Kaggle} & 
\href{https://www.kaggle.com/competitions/sberbank-russian-housing-market/rules\#7-competition-data}{Subject to Competition Rules} \\

\midrule
\multicolumn{3}{l}{\textbf{\textit{TimeSeriesGym Originals}}} \\
\midrule

\begin{tabular}[c]{@{}l@{}}Convert ResNet TensorFlow\\ implementation to PyTorch\end{tabular} & 
\href{https://github.com/hfawaz/dl-4-tsc/blob/master/classifiers/resnet.py}{GitHub} & 
\href{https://github.com/hfawaz/dl-4-tsc/blob/master/LICENSE}{GNU General Public License v3.0} \\

\begin{tabular}[c]{@{}l@{}}Convert STOMP Algorithm\\ Implementation in R to Python\end{tabular} & 
\href{https://github.com/matrix-profile-foundation/tsmp}{GitHub} & 
\href{https://github.com/matrix-profile-foundation/tsmp/blob/master/LICENSE.md}{Apache License 2.0} \\

\begin{tabular}[c]{@{}l@{}}Evaluate MOIRAI time series\\ foundation model on the\\ Context Is Key (CiK) benchmark\end{tabular} & 
\href{https://github.com/ServiceNow/context-is-key-forecasting}{GitHub} & 
\href{https://github.com/ServiceNow/context-is-key-forecasting/blob/main/LICENSE.md}{Apache License 2.0} \\

\begin{tabular}[c]{@{}l@{}}Evaluate Chronos time series\\ foundation model on the NN5 dataset\\ within Context Is Key (CiK) benchmark\end{tabular} & 
-- & 
\href{https://github.com/ServiceNow/context-is-key-forecasting/blob/main/LICENSE.md}{Apache License 2.0} \\

\begin{tabular}[c]{@{}l@{}}GIFT-EVAL: A Benchmark for General\\ Time Series Forecasting Model Evaluation\end{tabular} & 
\href{https://github.com/SalesforceAIResearch/gift-eval}{GitHub} & 
\href{https://github.com/SalesforceAIResearch/gift-eval/blob/main/LICENSE.txt}{Apache License 2.0} \\

\begin{tabular}[c]{@{}l@{}}Hexagon ML UCR Time Series\\ Anomaly Detection\end{tabular} & 
\href{https://www.cs.ucr.edu/~eamonn/time_series_data_2018/}{UCR} & 
Not available \\

\begin{tabular}[c]{@{}l@{}}Long Horizon Time Series Forecasting\\ Using Time Series Library\end{tabular} & 
\href{https://github.com/thuml/Time-Series-Library}{GitHub} & 
\href{https://github.com/thuml/Time-Series-Library/blob/main/LICENSE}{MIT License} \\

\begin{tabular}[c]{@{}l@{}}Long-Horizon Weather Forecasting\\ using Time Series Library's Itransformer\end{tabular} & 
-- & 
\href{https://github.com/thuml/Time-Series-Library/blob/main/LICENSE}{MIT License} \\

MIT-BIH ECG Arrhythmia Detection & 
\href{https://www.physionet.org/content/mitdb/1.0.0/}{PhysioNet} & 
\href{https://www.physionet.org/content/mitdb/view-license/1.0.0/}{Open Data Commons Attribution License v1.0} \\

\begin{tabular}[c]{@{}r@{}}MOMENT for Anomaly Detection\\ on UCR datasets\end{tabular} & 
\href{https://github.com/moment-timeseries-foundation-model/moment}{GitHub} & 
\href{https://github.com/moment-timeseries-foundation-model/moment/blob/main/LICENSE}{MIT License} \\

PTB-XL ECG Classification & 
\href{https://physionet.org/content/ptb-xl/1.0.3/}{PhysioNet} & 
\href{https://physionet.org/content/ptb-xl/view-license/1.0.3/}{Creative Commons Attribution 4.0 International Public License} \\

\begin{tabular}[c]{@{}l@{}}TimeSeriesExam: A Time Series\\ Understanding Exam\end{tabular} & 
\href{https://huggingface.co/datasets/AutonLab/TimeSeriesExam1}{Hugging Face} & 
\href{https://github.com/moment-timeseries-foundation-model/TimeSeriesExam/blob/main/LICENSE}{MIT License} \\

\bottomrule
\end{tabular}%
}
\caption{This table provides transparency about the source and licensing information for each challenge in the \TimeSeriesGym benchmark. For the Kaggle challenges, most are subject to Kaggle's competition rules, with a few under Creative Commons licenses. The \TimeSeriesGym Original challenges come from diverse sources including GitHub repositories, HuggingFace, etc. with various open-source licenses (Apache, MIT, GPL, Creative Commons). This diversity of sources and licenses demonstrates the benchmark's foundation in accessible, reusable datasets and code while ensuring proper attribution and compliance with original creators' terms.}
\label{tab:time_series_sources}
\end{table}

\begin{table}[ht]
\centering
\caption{\TimeSeriesGym\texttt{-Lite} is a streamlined collection of six diverse time series challenges carefully selected to evaluate AI agents while balancing thoroughness with efficiency. The challenges cover essential ML engineering skills including basic data science, handling missing/multi-source data, code migration, hyperparameter optimization, research code modeling, and data labeling. The collection spans multiple domains (healthcare, finance, algorithms) and various time series tasks (classification, forecasting, anomaly detection, code migration). This cost-effective subset allows researchers to quickly benchmark agent capabilities across critical ML engineering skills without the resource requirements of the full \TimeSeriesGym benchmark.}
\label{tab:timeseriesgym_challenges_lite}
\resizebox{\textwidth}{!}{ 
\renewcommand{\arraystretch}{1.3} 
\begin{tabular}{r|ccc}
\toprule
\textbf{Challenge} & \textbf{Required Skills} & \textbf{Time Series Task} & \textbf{Domain} \\
\midrule
Child Mind Institute - & Basic data science & \multirow{2}{*}{Classification} & \multirow{2}{*}{Healthcare} \\
Detect Sleep States & (data handling and modeling) & & \\
Optiver Realized & Handling missing and & \multirow{2}{*}{Forecasting} & \multirow{2}{*}{Finance} \\
Volatility Prediction & multi-source data & & \\
Convert ResNet TensorFlow & \multirow{2}{*}{Classification} & \multirow{2}{*}{Code Migration} & \multirow{2}{*}{Algorithm} \\
implementation to PyTorch & & & \\
PTB-XL ECG & Hyperparameter optimization & \multirow{2}{*}{Classification} & \multirow{2}{*}{Healthcare} \\
Classification & \& model selection & & \\
MOMENT Anomaly & Modeling & \multirow{2}{*}{Anomaly Detection} & Healthcare, Gait, \\
Score Calculation & (Using research code) & & Synthetic, Energy, Devices \\
MIT-BIH Arrhythmia & \multirow{2}{*}{Data labeling} & \multirow{2}{*}{Classification} & \multirow{2}{*}{Healthcare} \\
Detection & & & \\
\bottomrule
\end{tabular}
}
\end{table}

\begin{table}[htbp]
\centering
\setlength{\tabcolsep}{6pt}
\renewcommand{\arraystretch}{1.2}
\caption{Average cost to run experiments on a single seed in the default evaluation setup \textit{i.e.} \texttt{AIDE} with \texttt{gpt-4.1-2025-04-14} with a maximum of 4 hours and 50 steps.}
\label{tab:average_costs}
\begin{tabular}{rl}
\toprule
\textbf{Benchmark}       & \textbf{Averge Cost (USD)} \\
\midrule
\TimeSeriesGym & 62.12 \\
\TimeSeriesGym\texttt{-Lite} & 7.96 \\
\bottomrule
\end{tabular}
\end{table}

\clearpage
\section{Implementation Details for Scaffolds}
\label{app:implementation_details_of_scaffolds}

\begin{table}[h!]
\centering
\caption{Scaffold hyperparameters. \texttt{\$TARGET\_MODEL} denotes the model being evaluated.}
\label{tab:scaffold-hyperparams}
\begin{tabular}{@{}ll@{}}
\toprule
\multicolumn{2}{l}{\texttt{AIDE}} \\
\midrule
\texttt{agent.code.model} & \texttt{\$TARGET\_MODEL} \\
\texttt{agent.feedback.model} & \texttt{gpt-4.1-2025-04-14} \\
\texttt{agent.steps} & \texttt{50} \\
\texttt{agent.search.max\_debug\_depth} & \texttt{20} \\
\texttt{agent.search.debug\_prob} & \texttt{1} \\
\texttt{agent.time\_limit} & \texttt{14400} \\
\texttt{exec.timeout} & \texttt{32400} \\
\midrule
\multicolumn{2}{l}{\texttt{OpenHands}} \\
\midrule
\texttt{agent} & \texttt{CodeActAgent} \\
\texttt{model} & \texttt{\$TARGET\_MODEL} \\
\texttt{max\_time\_in\_hours} & \texttt{4} \\
\texttt{max\_steps} & \texttt{50} \\
\bottomrule
\end{tabular}
\end{table}

Table~\ref{tab:scaffold-hyperparams} summarizes the hyperparameters used for the two scaffolds: \texttt{AIDE}~\cite{jiang2025aide} and \texttt{OpenHands}' \texttt{CodeAct}~\cite{wang2024openhands}.
 We did not directly use the MLE-bench~\cite{chan2025mlebench} modifications to the agentic scaffolds for two main reasons. First, the official scaffold implementation has undergone updates since MLE-bench. Second, certain changes made in MLE-bench were not applicable to our benchmark, which involves more diverse modes of evaluation. Therefore, we modified the latest version of the agentic scaffolds to enhance robustness, improve execution stability, and support a broader range of competitions.

\subsection{AIDE}

We forked the original \AIDE repository in May 2025 and added useful changes from the MLE-bench project. These include better API calls and support for more API providers. Our key modifications are summarized below:

\begin{enumerate}
    \item Updated instructions to cover all types of tasks, not just Kaggle challenges
    \item Modified the prompt to work with different file types, since our tasks use many input/output formats
    \item Fixed the interpreter handling to prevent timeouts and system hangs
\end{enumerate}

\subsection{OpenHands}

We forked the original \texttt{OpenHands} repository from tag \texttt{v0.34.0} (May 2025). We reduced the RAM allowance to 10 GiB (from 100 GiB) as we did not observe any memory-related issues during our tests.

\begin{figure}
    \caption{The prompt we use to initialize all scaffolds, adapted from \texttt{MLE-bench} \cite{chan2025mlebench}.}
    \label{fig:overall_instructions}
    \centering
    \begin{minted}[frame=single,fontsize=\small, breaklines=true]{text}
# Machine Learning Engineer Challenge: TimeSeriesGym Competition

## Your Role
You are a Machine Learning Engineer tasked with solving complex ML problems in the TimeSeriesGym offline competition. This competition tests your ability to understand problems, follow instructions, and build/train/evaluate ML models.

## Core Requirements
- You must solve each task by training a model and generating predictions on the test set
- Hand-labeling the data to produce submissions is strictly prohibited

## Environment Setup

### Critical Files and Locations
- **Competition Instructions**: Located at `/home/data/description.md` - This is your primary reference for understanding the specific problem
- **Dataset**: All data files are in `/home/data/` directory
- **Submission Path**: Your final submission MUST be saved to `/home/submission/` in the format specified in the instructions
- **Environment Instructions**: These general instructions can be found at `/home/instructions.txt`

### Test Dataset Information
- We've created our own train/test splits for evaluation purposes
- The data in `/home/data/` may differ slightly from what's described in the competition instructions
- Use the provided train/test splits for your work

## Submission Instructions

### File Format and Location
- Your submission MUST be saved to the exact path: `/home/submission/`
- The file format must strictly follow the specifications in `description.md`
- Failing to use the correct path or format will result in a failed evaluation

### Validation Process
1. Before finalizing your submission, validate its format using the validation server
2. Use the command: `curl -X POST -F "file=@${SUBMISSION_FILE}" http://localhost:5000/validate`
3. Alternatively, use the provided script: `/home/validate_submission.sh`
4. The validator will confirm if your submission format is valid (but won't provide a score)
5. Make corrections if needed and revalidate until you receive a confirmation

### Final Submission
- Once validated, ensure your final submission file remains at `/home/submission/`
- Only the last file in this location will be evaluated
- Double-check that your model's predictions cover the entire test set
- Review the submission format requirements one last time before completing

## Evaluation Criteria
- Your submission will be graded based on the metric specified in the competition instructions
- Your goal is to achieve the highest possible score

## Important Rules
- Viewing or copying others' solutions is prohibited and will result in disqualification
- In case of conflicts between these instructions and `/home/data/description.md`, these general instructions take priority
\end{minted}    
\end{figure}

\clearpage
\section{Detailed Evaluation Results}
\subsection{Full Benchmark Evaluation Result}
\begin{table}[htbp]
\centering
\renewcommand{\arraystretch}{1.15} 
\resizebox{0.75\textwidth}{!}{%
\begin{tabular}{r|cccc}
\toprule
\textbf{Challenge} & \textbf{Evaluation Metric} & \textbf{Best @ 3} & \textbf{Average @ 3}\\
\midrule
\multicolumn{3}{l}{\textit{\textbf{Kaggle Challenges}}} \\
\midrule
\begin{tabular}[c]{@{}c@{}}AMP-Parkinson's Disease\\ Progression Prediction\end{tabular} & \begin{tabular}[c]{@{}c@{}}Symmetric Mean Absolute\\ Percentage Error\end{tabular} & 111.22 &	120.50\\
ASHRAE - Great Energy Predictor III  & \begin{tabular}[c]{@{}c@{}}Root Mean Square\\ Logarithmic Error\end{tabular} & 1.02 & 1.92\\
Child Mind Institute-- Detect Sleep States & \begin{tabular}[c]{@{}c@{}}Event Detection\\ Average Precision\end{tabular} & 0.02 & 0.01\\
Google Brain - Ventilator Pressure Prediction & Mean Absolute Error & 0.58 & 5.40\\
G2Net Gravitational Wave Detection & \begin{tabular}[c]{@{}c@{}}Area Under\\ ROC Curve\end{tabular} & 0.51 & 0.50\\
HMS - Harmful Brain Activity Classification & KL Divergence & 1.16 &	1.56\\
LANL Earthquake Prediction & Mean Absolute Error & 2.18 &	2.89\\
M5 Forecasting - Accuracy & \begin{tabular}[c]{@{}c@{}}Weighted Root Mean Squared\\ Scaled Error\end{tabular} & 0.82 & 3.13\\
Online Product Sales & \begin{tabular}[c]{@{}c@{}}Root Mean Square\\ Logarithmic Error\end{tabular} & 0.91 & 1.08\\
Optiver Realized Volatility Prediction & \begin{tabular}[c]{@{}c@{}}Root Mean Square\\ Percentage Error\end{tabular} & 0.28 & 0.30\\
Recruit Restaurant Visitor Forecasting & \begin{tabular}[c]{@{}c@{}}Root Mean Square\\ Logarithmic Error\end{tabular} & 0.55 & 0.60\\
Sberbank Russian Housing Market & \begin{tabular}[c]{@{}c@{}}Root Mean Square\\ Logarithmic Error\end{tabular} & 0.39 & 0.40\\
\midrule
\multicolumn{3}{l}{\textit{\textbf{TimeSeriesGym Originals}}} \\
\midrule
\begin{tabular}[c]{@{}r@{}}Convert ResNet TensorFlow\\implementation to PyTorch\end{tabular} & Custom Code Grading Test Cases & 5/9	& 5/9\\
\begin{tabular}[c]{@{}r@{}}Convert STOMP Algorithm\\implementation in R to Python\end{tabular} & Custom Code Grading Test Cases &  2/4	& 1.6/4\\
\begin{tabular}[c]{@{}r@{}}Evaluate MOIRAI time series foundation model\\ on the Context Is Key (CiK) benchmark\end{tabular} & Resolved (Binary) & N/A & N/A \\
\begin{tabular}[c]{@{}r@{}}Evaluate Chronos time series foundation model\\ on the NN5 dataset within Context Is Key\\ (CiK) benchmark\end{tabular} & Resolved (Binary) & N/A & N/A \\
Implement \& Evaluate CSDI to Impute PM2.5 Data & Mean Absolute Error & N/A & N/A \\ \\
Train \& Evaluate CSDI to Impute PM2.5 Data & Mean Absolute Error & N/A & N/A \\
\begin{tabular}[c]{@{}r@{}}GIFT-EVAL: A Benchmark for General Time Series*\\ Forecasting Model Evaluation\end{tabular} & Mean Absolute Percentage Error & N/A & N/A \\
Hexagon ML UCR Time Series Anomaly Detection* & \begin{tabular}[c]{@{}c@{}}Adjusted Best\\ F1 Score\end{tabular} & 0.38 & 0.38\\
\begin{tabular}[c]{@{}r@{}}Long Horizon Time Series Forecasting\\ Using Time Series Library\end{tabular} & \begin{tabular}[c]{@{}c@{}}Mean Squarred\\ Error\end{tabular} & N/A & N/A\\
\begin{tabular}[c]{@{}r@{}}Long-Horizon Weather Forecasting using\\ Time Series Library's Itransformer\end{tabular} & Exact Match (Binary) & N/A & N/A \\
MIT-BIH ECG Arrhythmia Detection & Accuracy & 0.87 & 0.84\\
\begin{tabular}[c]{@{}r@{}}MOMENT for Anomaly Detection\\ on UCR datasets\end{tabular} & Exact Match (Binary) & N/A & N/A\\
PTB-XL ECG Classification & Accuracy & 0.81 & 0.80\\
TimeSeriesExam: A Time Series Understanding Exam & Accuracy & N/A & N/A\\
\midrule
\multicolumn{3}{l}{\textit{\textbf{Derived Challenges}}} \\
\midrule
Google Brain - Ventilator Pressure Prediction\\With Missingness & \begin{tabular}[c]{@{}c@{}}Mean Absolute\\ Error\end{tabular} & 2.72 & 6.66\\
Improve PTB-XL ECG Classification Code & \begin{tabular}[c]{@{}c@{}}Code Enhancement\\(Experiment Tracking,\\Readability, Reproducibility)\end{tabular} & N/A & N/A\\
\begin{tabular}[c]{@{}r@{}}MIT-BIH Arrhythmia Detection\\with Weak Supervision\end{tabular} & Accuracy & 0.87 & 0.77 \\
Optiver Realized Volatility Prediction\\With Missingness & \begin{tabular}[c]{@{}c@{}}Root Mean Square\\ Percentage Error\end{tabular} & 0.33 & 0.33\\
\begin{tabular}[c]{@{}r@{}}Optiver Realized Volatility Prediction\\ with Hyper-parameter Optimization\end{tabular} & \begin{tabular}[c]{@{}c@{}}Improvement in Root Mean\\ Square Percentage Error\end{tabular} & -0.01&-0.15\\
\begin{tabular}[c]{@{}r@{}}PTB-XL ECG Classification\\ with Hyperparameter Optimization\end{tabular} & Improvement in Accuracy & 0.08 & 0.03\\
\bottomrule
\end{tabular}%
}
\caption{Comprehensive performance metrics for AI agents on all \TimeSeriesGym challenges, including best and average scores from three runs. Agents struggle to solve \TimeSeriesGym Original challenges. Derived challenges demonstrate how added complexity (missingness, hyperparameter optimization) affects performance. These results highlight both the capabilities and limitations of current ML engineering agents across diverse time series tasks.}
\label{tab:main_result_raw}
\end{table}

We provide detailed evaluation results for each task in \TimeSeriesGym in Table~\ref{tab:main_result_raw}. Each task was executed with three random seeds; we report both the average and best scores across these runs. Entries marked \texttt{N/A} indicate that the agent failed to produce a valid solution due to exceeding the time- or step-limit. For the \texttt{GIFT-Eval} and \texttt{UCR Anomaly Detection} challenges, evaluation is performed on a subset of the original benchmark, since our focus is on assessing the agent’s ability to leverage the research repository rather than full benchmark performance.

\subsection{Ablation Study Evaluation Result}
\begin{table}[htbp]
\centering
\scriptsize
\setlength{\tabcolsep}{4pt}
\renewcommand{\arraystretch}{1.2}
\begin{tabular}{@{}p{3cm}*{6}{>{\centering\arraybackslash}p{1.3cm}}@{}}
\toprule
\textbf{Challenge}
  & \textbf{8 hours / 100 steps}
  & \textbf{12 hours / 150 steps}
  & \textbf{OpenHands}
  & \textbf{o3}
  & \textbf{claude-3-7}
  & \textbf{No Reminder} \\
\midrule
\begin{tabular}[c]{@{}l@{}}Child Mind Institute––\\ Detect Sleep States\end{tabular}
  & 0.02 / 0.02 & 0.00 / 0.00 & 0.00 / 0.00 & 0.11 / 0.11 & \texttt{N/A}    & \texttt{N/A}    \\

\begin{tabular}[c]{@{}l@{}}Optiver Realized Volatility\\ Prediction with Missingness\end{tabular}
  & 0.32 / 0.33 & 0.31 / 0.31 & 0.64 / 0.64 & 0.42 / 0.43 & 0.25 / 0.25 & 0.32 / 0.32 \\

Convert ResNet TensorFlow to PyTorch
  & 0.56 / 0.56         & 0.89 / 0.89         & 0.56 / 0.44       & 0.56 / 0.56         & 0.89 / 0.78        & 0.56 / 0.56         \\

\begin{tabular}[c]{@{}l@{}}PTB-XL ECG Classification with\\ Hyperparameter Search\end{tabular}
  & 0.45 / 0.22 & 0.45 / 0.10 & \texttt{N/A} & 0.14 / 0.10 & 0.09 / 0.06 & 0.05 / 0.03 \\

MOMENT Anomaly Score Calculation
  & \texttt{N/A} & \texttt{N/A} & 0.00 / 0.00 & 0.00 / 0.00 & \texttt{N/A} & \texttt{N/A} \\

\begin{tabular}[c]{@{}l@{}}MIT-BIH Arrhythmia Detection\\ with Weak Supervision\end{tabular}
  & 0.83 / 0.56 & 0.80 / 0.60 & 0.73 / 0.72 & 0.53 / 0.45 & 0.79 / 0.66 & 0.74 / 0.68 \\
\bottomrule
\end{tabular}
\caption{This table presents detailed ablation study results comparing agent performance across six different configurations on the TimeSeriesGym-Lite benchmark. Each cell shows Best@3/Avg@3 scores, with N/A indicating no valid solutions. The experiments compare time variations (8 hours/100 steps vs. 12 hours/150 steps), scaffold differences (OpenHands), model types (o3, claude-3-7), and whether agents are reminded of remaining time. Results show mixed effects of increased time allocation, with certain challenges (ResNet conversion) benefiting significantly while others show minimal improvements or even degradation. Both model type and scaffold selection substantially impact performance, with different models excelling on different challenges. This highlights the complex interplay between agent configurations and task types in ML engineering.}
\label{tab:ablation}
\end{table}

\clearpage
\section{Failure Mode Illustration}
\label{app:failure_mode}

\subsection{Agents Miss Important Information}

\begin{figure}
\centering
\begin{minted}[frame=single,fontsize=\small, breaklines=true]{python}
step 19
FileReadAction(path='/home/data/moment/momentfm/models/moment.py', 
start=0, end=-1, thought='', action=<ActionType.READ: 'read'>, 
security_risk=None, impl_source=<FileReadSource.OH_ACI: 'oh_aci'>, 
view_range=[1, 60])

step 20
FileReadAction(path='/home/data/moment/momentfm/models/moment.py', 
start=0, end=-1, thought='', action=<ActionType.READ: 'read'>, 
security_risk=None, impl_source=<FileReadSource.OH_ACI: 'oh_aci'>, 
view_range=[61, 120])

...

step 25
FileReadAction(path='/home/data/moment/momentfm/models/moment.py', 
start=0, end=-1, thought='', action=<ActionType.READ: 'read'>, 
security_risk=None, impl_source=<FileReadSource.OH_ACI: 'oh_aci'>, 
view_range=[361, 420]
\end{minted}
\caption{\texttt{OpenHands} wastes 5 steps on inspecting model file while the correct way to import the model is in README}
\label{lst:openhands-load-moment}
\end{figure}

As illustrated in Listing~\ref{lst:openhands-load-moment}, \OpenHands spends five consecutive steps scanning to the end of the model file in an attempt to discover the correct import method for \texttt{MOMENT}. This behavior reveals two critical shortcomings. First, the agent follows a greedy, linear scanning strategy with no early‐stop criterion or hierarchical search plan—it blindly paginates through the file rather than formulating a focused query. Second, it fails to leverage the README, which explicitly documents the proper import instructions. Together, these issues demonstrate a lack of strategic planning and contextual awareness. We observed a similar pattern in \AIDE, where blind iteration and omission of available documentation likewise impede efficient problem solving.

\subsection{\AIDE Interpreter Execution Can Trigger Undesirable Behavior}

\begin{figure}
\centering
\begin{minted}[frame=single,fontsize=\small, breaklines=true]{python}
global_scope: dict = {}
while True:
    code = code_inq.get()
    os.chdir(str(self.working_dir))
    with open(self.agent_file_name, "w") as f:
        f.write(code)

    event_outq.put(("state:ready",))
    try:
        exec(compile(code, self.agent_file_name, "exec"), global_scope)
    except BaseException as e:
        ...
\end{minted}
\caption{\AIDE’s interpreter does not execute code under main environment}
\label{lst:aide-interpreter}
\end{figure}

\noindent
Shown in Listing~\ref{lst:aide-interpreter}, \AIDE invokes Python’s \texttt{exec} in a persistent \texttt{global\_scope}, then employs an LLM‐based “judge” to inspect the generated code and its stdout. Any logic guarded by
  \texttt{if \_\_name\_\_ == \texttt{"\_\_main\_\_"}:}
will be skipped—because \texttt{global\_scope} does not set \_\_name\_\_ to \texttt{"\_\_main\_\_"}. As a result, the judge may erroneously declare such runs valid, even when critical execution paths never occur, and further retries or debug steps cannot correct this oversight.

\subsection{\texttt{AIDE}  Single‐File Approach is Error-prone}
\begin{figure}
\centering
\begin{minted}[frame=single,fontsize=\small, breaklines=true]{python}
import os
import subprocess
import sys
import shutil
import numpy as np

def install_requirements(tsl_dir):
    req_file = os.path.join(tsl_dir, "requirements.txt")
    req_file_abs = os.path.abspath(req_file)
    print(f"Installing requirements from {req_file_abs} ...")
    try:
        subprocess.run(
            [sys.executable, "-m", "pip", "install", "--upgrade", "pip"], check=True
        )
        subprocess.run([sys.executable, "-m", "pip", "install", "wheel"], check=True)
        subprocess.run(
            [sys.executable, "-m", "pip", "install", "-r", req_file_abs],
            check=True,
            cwd=tsl_dir,
        )
    except subprocess.CalledProcessError as e:
        print("Failed to install requirements!")
        print("Output:", e.output if hasattr(e, "output") else "No output")
        sys.exit(1)

def prepare_weather_data(tsl_dir, input_dir):
    dataset_dir = os.path.join(tsl_dir, "dataset", "weather")
    os.makedirs(dataset_dir, exist_ok=True)
    src_weather = os.path.join(input_dir, "weather.csv")
    dst_weather = os.path.join(dataset_dir, "weather.csv")
    if not os.path.exists(dst_weather):
        print(f"Copying {src_weather} to {dst_weather}")
        shutil.copy(src_weather, dst_weather)
    else:
        print(f"{dst_weather} already exists.")

def run_itransformer(tsl_dir, submission_dir):
    runpy = os.path.join(tsl_dir, "run.py")
    pred_file = os.path.abspath(os.path.join(submission_dir, "pred.npy"))
    # Set label_len to 96 as per the competition instructions
    cmd = [
        sys.executable,
        runpy,
        "--task_name",
        "long_term_forecast",
    ... #skip to save space

if __name__ == "__main__":
    main()
\end{minted}
\caption{\AIDE's solution for utilizing \texttt{Time-Series-Library} repository}
\label{lst:aide-tsl-forecast}
\end{figure}

As shown in Listing~\ref{lst:aide-tsl-forecast}, \AIDE encapsulates the entire forecasting workflow in a single script. Whenever it must invoke system commands, it relies on Python’s \texttt{subprocess} module—an approach that can obscure full tracebacks and miss intermediate errors. Furthermore, to import modules from the research repository, \AIDE repeatedly alters the Python search path or changes the working directory (e.g., via \texttt{sys.path.append}), which is inefficient and brittle.

\clearpage
\section{Two-Faceted Grading Approach}
\label{app:holistic_grading}

We implemented a dual grading methodology to provide comprehensive evaluation of submitted solutions. This section introduces the general concept of our two complementary evaluation approaches: exact grading and judge-based grading.

\subsection{Grading Methodologies}

\subsubsection{Exact Grading Methodology}
The exact grading approach represents a deterministic, checklist-based evaluation focused on verifying specific required components. This objective method evaluates submissions against explicit criteria with binary pass/fail outcomes, providing clear feedback on technical requirements. The exact grading methodology emphasizes quantifiable metrics and compliance with predefined specifications.

Key aspects of exact grading include:
\begin{itemize}
    \item Binary verification of required components (present/absent)
    \item Point-by-point scoring against a predefined checklist
    \item Focus on technical compliance with specifications
    \item Reproducible results with minimal subjective interpretation
\end{itemize}

\subsubsection{Judge-Based Qualitative Methodology}
The judge-based approach provides a nuanced evaluation that assesses artifacts beyond mere presence of required components. This method employs large language models (LLMs) as judges to evaluate submissions against custom criteria with chain-of-thought reasoning.

Key aspects of judge grading include:
\begin{itemize}
    \item Scoring on a continuous scale
    \item Evaluation of code quality, architecture design, and implementation elegance
    \item Detailed reasoning explaining score justification
    \item Ability to recognize exceptional implementations that exceed basic requirements
\end{itemize}

\subsection{Implementation for PTB-XL Classification Challenge}

\subsubsection{Exact Grading Implementation}
For the PTB-XL Classification Challenge, our exact grading implementation evaluates code artifact submissions through:

\begin{enumerate}
    \item \textbf{Feature Extraction}: Using regular expression pattern matching and AST parsing to identify required code components.
    \item \textbf{Binary Verification}: Checking each requirement against pass/fail criteria.
    \item \textbf{Static Analysis}: Using linting tools to check against PEP 8 standards.
    \item \textbf{File Structure Validation}: Verifying required files and directories.
\end{enumerate}

The exact grading for this challenge evaluates four primary categories, each worth 25\% of the final score:
\begin{itemize}
    \item \textbf{TensorBoard Usage}: Proper imports, SummaryWriter initialization, metric logging, etc.
    \item \textbf{Code Quality}: Syntax verification, docstrings, type annotations, and PEP 8 compliance.
    \item \textbf{Hydra Configuration}: Proper imports, decorator usage, and configuration files.
    \item \textbf{Model Accuracy}: Prediction accuracy against ground truth labels.
\end{itemize}

\subsubsection{Judge-Based Implementation}
For this challenge, we employed G-Eval~\cite{liu2023g}, a framework that uses LLMs with chain-of-thought reasoning. The implementation evaluates code through:

\begin{enumerate}
    \item \textbf{Evaluation Steps}: Using predefined steps for chain-of-thought reasoning.
    \item \textbf{Comprehensive Assessment}: Evaluating multiple parameters including code structure and architecture decisions.
    \item \textbf{Score Calculation}: Generating normalized scores on a 0.0-1.0 scale.
    \item \textbf{Reasoning Provision}: Providing detailed explanations for the evaluation.
\end{enumerate}

\subsection{Comparative Analysis}

The two approaches serve complementary purposes:

\begin{table}[h]
  \centering
  \resizebox{\textwidth}{!}{%
    \begin{tabular}{r|c|c}
      \toprule
      \textbf{Aspect} & \textbf{Exact Grading} & \textbf{Judge Grading} \\
      \midrule
      Objectivity 
        & High (deterministic) 
        & Moderate (LLM-based) \\
        
      Granularity 
        & Binary (present/absent) 
        & Continuous (quality scores) \\
        
      Feedback Detail 
        & Limited (requirement verification) 
        & Rich (explanatory reasoning) \\
        
      Reproducibility 
        & High (automated) 
        & Moderate (LLM consistency) \\
        
      Evaluation Scope 
        & Technical compliance 
        & Code quality, effectiveness \\
        
      Methodology 
        & Rule-based checks 
        & LLM with chain-of-thought \\
        
      Scalability 
        & Low (manual rule design) 
        & High (natural language criteria) \\
      \bottomrule
    \end{tabular}%
  }
  \caption{Transposed Comparison of Exact and Judge-Based Grading Approaches}
  \label{tab:grading_comparison_transposed}
\end{table}

\subsection{Combined Grading Benefits}

Using both approaches provides several advantages:
\begin{itemize}
    \item Ensures baseline technical requirements are met (exact grading)
    \item Rewards exceptional implementations and identifies subtle weaknesses (judge grading)
    \item Balances objective verification with subjective quality assessment
    \item Provides comprehensive feedback on both technical compliance and code quality
    \item Creates a fair and holistic evaluation system
\end{itemize}

\subsection{Grading Examples for PTB-XL Challenge}

Below are example outputs from both grading systems applied to the same submission for the PTB-XL Classification Challenge.

\subsubsection{Exact Grading Output}
\begin{minted}[frame=single,fontsize=\small, breaklines=true]{text}
TensorBoard Usage (25% of total score)
TensorBoard SummaryWriter is properly imported: 0.2/0.2
SummaryWriter is initialized: 0.2/0.2
Metrics are logged with add_scalar: 0.2/0.2
Model graph is logged with add_graph: 0.2/0.2
Found 2 log files in runs/
TensorBoard log files are present: 0.2/0.2
Final score: 1.00/1.00

Code Quality and Documentation (25% of total score)
Syntax check passed: 0.1/0.1
Model file has module docstring: 0.1/0.1
Training file has module docstring: 0.1/0.1
Both files have type annotations: 0.1/0.1
Model classes have docstrings: 0.05/0.05
Functions have docstrings: 0.1/0.1
Docstrings have Args/Returns sections: 0.1/0.1
Linting score: 0.25/0.25
Model file has no PEP 8 violations
Training file has no PEP 8 violations
Model file follows style guidelines
Training file follows style guidelines
Final score: 0.90/0.90

Hydra Configuration Usage (25% of total score)
Hydra is properly imported: 0.2/0.2
@hydra.main decorator is used: 0.2/0.2
OmegaConf/DictConfig is used: 0.2/0.2
Config is used for model parameters: 0.2/0.2
Config file exists with model parameters: 0.2/0.2
Final score: 1.00/1.00

Model Accuracy (25% of total score)
Model prediction accuracy: 1.0/1.0

Evaluation Summary
TensorBoard score: 1.00/1.0 (25% weight)
Code quality score: 1.00/1.0 (25% weight)
Hydra config score: 1.00/1.0 (25% weight)
Model accuracy: 1.0 (25% weight)
Overall score: 1.00/1.0
\end{minted}

\subsubsection{Judge-Based Grading Output}

\begin{minted}[frame=single,fontsize=\small, breaklines=true]{text}
File: example/model.py
Type: Model Script

- Code Quality and Documentation
  Score: 0.90
  Reason: The module has clear docstrings explaining the model's purpose and architecture. Function parameters and return types are well-annotated. Class and method docstrings include accurate Args and Returns sections. The code adheres to PEP 8, with proper spacing and naming conventions. The architecture is logically structured, but the module-level docstring could be more detailed.

- Model Architecture Design
  Score: 0.93
  Reason: The model utilizes configuration parameters effectively. Architecture includes convolutional layers suitable for ECG classification. Implements an efficient forward method and utility functions like parameter counting. Supports hyperparameter flexibility. Minor issue: model summary function could be better integrated.

- Model Configuration Handling
  Score: 0.86
  Reason: Configuration object is accepted with fallback defaults. Parameters are correctly extracted from config. Compatible with Hydra; well-documented parameter usage. Lacks explicit demonstration of usage with multiple configurations.

------------------------------------------------------

File: example/train.py
Type: Training Script

- TensorBoard Usage
  Score: 1.00
  Reason: SummaryWriter is correctly imported and initialized. Metrics are logged with add_scalar. Model graph is logged with add_graph. Writer is closed properly after training.

- Code Quality and Documentation
  Score: 0.93
  Reason: Clear module-level docstring and good use of type annotations. Functions are well-documented with Args and Returns. Adheres to PEP 8. Code structure is logical, variable naming is clear. Minor improvements possible in consistency.

- Hydra Configuration Usage
  Score: 1.00
  Reason: Hydra is imported and used with @hydra.main. OmegaConf and DictConfig are correctly used. Configuration passed to model with appropriate config_path/config_name.

- Model Training Completeness
  Score: 0.96
  Reason: Includes full training pipeline: data loading, preprocessing, training/validation loops. Implements loss calculation, optimizer, LR scheduling, checkpointing, and final predictions.
\end{minted}

%% file: neurips_2025.bbl
\begin{thebibliography}{10}

\bibitem{bogin-etal-2024-super}
Ben Bogin, Kejuan Yang, Shashank Gupta, Kyle Richardson, Erin Bransom, Peter Clark, Ashish Sabharwal, and Tushar Khot.
\newblock {SUPER}: Evaluating agents on setting up and executing tasks from research repositories.
\newblock In Yaser Al-Onaizan, Mohit Bansal, and Yun-Nung Chen, editors, {\em Proceedings of the 2024 Conference on Empirical Methods in Natural Language Processing}, pages 12622--12645, Miami, Florida, USA, November 2024. Association for Computational Linguistics.

\bibitem{cai2024timeseriesexam}
Yifu Cai, Arjun Choudhry, Mononito Goswami, and Artur Dubrawski.
\newblock {TimeSeriesExam}: A time series understanding exam.
\newblock {\em NeurIPS'24 Time Series in the Age of Large Models Workshop}, 2024.

\bibitem{chan2025mlebench}
Jun~Shern Chan, Neil Chowdhury, Oliver Jaffe, James Aung, Dane Sherburn, Evan Mays, Giulio Starace, Kevin Liu, Leon Maksin, Tejal Patwardhan, Aleksander Madry, and Lilian Weng.
\newblock {MLE}-bench: Evaluating machine learning agents on machine learning engineering.
\newblock In {\em The Thirteenth International Conference on Learning Representations}, 2025.

\bibitem{embretson2013item}
Susan~E Embretson and Steven~P Reise.
\newblock {\em Item response theory for psychologists}.
\newblock Psychology Press, 2013.

\bibitem{goswami2023aqua}
Mononito Goswami, Vedant Sanil, Arjun Choudhry, Arvind Srinivasan, Chalisa Udompanyawit, and Artur Dubrawski.
\newblock {Aqua}: A benchmarking tool for label quality assessment.
\newblock {\em Advances in Neural Information Processing Systems}, 36:79792--79807, 2023.

\bibitem{goswami2024moment}
Mononito Goswami, Konrad Szafer, Arjun Choudhry, Yifu Cai, Shuo Li, and Artur Dubrawski.
\newblock {MOMENT}: A family of open time-series foundation models.
\newblock In {\em International Conference on Machine Learning}, pages 16115--16152. PMLR, 2024.

\bibitem{guinet2024automated}
Gauthier Guinet, Behrooz Omidvar-Tehrani, Anoop Deoras, and Laurent Callot.
\newblock Automated evaluation of retrieval-augmented language models with task-specific exam generation.
\newblock In {\em Forty-first International Conference on Machine Learning}, 2024.

\bibitem{ho2024map}
Cherie Ho, Jiaye Zou, Omar Alama, Sai~Mitheran Jagadesh~Kumar, Cheng-Yu Chiang, Taneesh Gupta, Chen Wang, Nikhil Keetha, Katia Sycara, and Sebastian Scherer.
\newblock Map it anywhere: Empowering bev map prediction using large-scale public datasets.
\newblock {\em Advances in Neural Information Processing Systems}, 37:64433--64453, 2024.

\bibitem{huang2024mlagentbench}
Qian Huang, Jian Vora, Percy Liang, and Jure Leskovec.
\newblock {MLA}gentbench: Evaluating language agents on machine learning experimentation.
\newblock In {\em Forty-first International Conference on Machine Learning}, 2024.

\bibitem{jain2024livecodebench}
Naman Jain, King Han, Alex Gu, Wen-Ding Li, Fanjia Yan, Tianjun Zhang, Sida Wang, Armando Solar-Lezama, Koushik Sen, and Ion Stoica.
\newblock Livecodebench: Holistic and contamination free evaluation of large language models for code.
\newblock {\em arXiv preprint arXiv:2403.07974}, 2024.

\bibitem{jiang2025aide}
Zhengyao Jiang, Dominik Schmidt, Dhruv Srikanth, Dixing Xu, Ian Kaplan, Deniss Jacenko, and Yuxiang Wu.
\newblock Aide: Ai-driven exploration in the space of code.
\newblock {\em arXiv preprint arXiv:2502.13138}, 2025.

\bibitem{jing2025dsbench}
Liqiang Jing, Zhehui Huang, Xiaoyang Wang, Wenlin Yao, Wenhao Yu, Kaixin Ma, Hongming Zhang, Xinya Du, and Dong Yu.
\newblock {DSB}ench: How far are data science agents from becoming data science experts?
\newblock In {\em The Thirteenth International Conference on Learning Representations}, 2025.

\bibitem{liang2022holistic}
Percy Liang, Rishi Bommasani, Tony Lee, Dimitris Tsipras, Dilara Soylu, Michihiro Yasunaga, Yian Zhang, Deepak Narayanan, Yuhuai Wu, Ananya Kumar, et~al.
\newblock Holistic evaluation of language models.
\newblock {\em arXiv preprint arXiv:2211.09110}, 2022.

\bibitem{liu2023g}
Yang Liu, Dan Iter, Yichong Xu, Shuohang Wang, Ruochen Xu, and Chenguang Zhu.
\newblock G-eval: Nlg evaluation using gpt-4 with better human alignment.
\newblock In {\em Proceedings of the 2023 Conference on Empirical Methods in Natural Language Processing}, pages 2511--2522, 2023.

\bibitem{massey1951kolmogorov}
Frank~J Massey~Jr.
\newblock The kolmogorov-smirnov test for goodness of fit.
\newblock {\em Journal of the American statistical Association}, 46(253):68--78, 1951.

\bibitem{mialon2023gaia}
Gr{\'e}goire Mialon, Cl{\'e}mentine Fourrier, Thomas Wolf, Yann LeCun, and Thomas Scialom.
\newblock {GAIA}: a benchmark for general ai assistants.
\newblock In {\em The Twelfth International Conference on Learning Representations}, 2023.

\bibitem{nathani2025mlgymnewframeworkbenchmark}
Deepak Nathani, Lovish Madaan, Nicholas Roberts, Nikolay Bashlykov, Ajay Menon, Vincent Moens, Amar Budhiraja, Despoina Magka, Vladislav Vorotilov, Gaurav Chaurasia, Dieuwke Hupkes, Ricardo~Silveira Cabral, Tatiana Shavrina, Jakob Foerster, Yoram Bachrach, William~Yang Wang, and Roberta Raileanu.
\newblock {MLG}ym: A new framework and benchmark for advancing ai research agents, 2025.

\bibitem{padigela2025ml}
Harshith Padigela, Chintan Shah, and Dinkar Juyal.
\newblock {ML-Dev-Bench}: Comparative analysis of ai agents on ml development workflows.
\newblock {\em arXiv preprint arXiv:2502.00964}, 2025.

\bibitem{metakaggle}
Megan Risdal and Timo Bozsolik.
\newblock Meta kaggle, 2022.

\bibitem{tang2023ml}
Xiangru Tang, Yuliang Liu, Zefan Cai, Yanjun Shao, Junjie Lu, Yichi Zhang, Zexuan Deng, Helan Hu, Kaikai An, Ruijun Huang, et~al.
\newblock {ML-B}ench: Evaluating large language models and agents for machine learning tasks on repository-level code.
\newblock {\em arXiv preprint arXiv:2311.09835}, 2023.

\bibitem{wang2024openhands}
Xingyao Wang, Boxuan Li, Yufan Song, Frank~F Xu, Xiangru Tang, Mingchen Zhuge, Jiayi Pan, Yueqi Song, Bowen Li, Jaskirat Singh, et~al.
\newblock Openhands: An open platform for ai software developers as generalist agents.
\newblock In {\em The Thirteenth International Conference on Learning Representations}, 2024.

\bibitem{wijk2024re}
Hjalmar Wijk, Tao Lin, Joel Becker, Sami Jawhar, Neev Parikh, Thomas Broadley, Lawrence Chan, Michael Chen, Josh Clymer, Jai Dhyani, et~al.
\newblock Re-bench: Evaluating frontier ai r\&d capabilities of language model agents against human experts.
\newblock {\em arXiv preprint arXiv:2411.15114}, 2024.

\bibitem{ye2024beyond}
Wen Ye, Yizhou Zhang, Wei Yang, Lumingyuan Tang, Defu Cao, Jie Cai, and Yan Liu.
\newblock Beyond forecasting: Compositional time series reasoning for end-to-end task execution.
\newblock {\em arXiv preprint arXiv:2410.04047}, 2024.

\end{thebibliography}
